\pdfoutput=1
\documentclass[dvipsnames,11pt]{article}

\usepackage[]{acl}

\usepackage{xcolor}
\usepackage{times}
\usepackage{hyperref}
\usepackage{latexsym}
\usepackage{circledsteps}
\usepackage{graphicx}
\usepackage{tikz}
\usepackage[T1]{fontenc}

\usepackage[algo2e]{algorithm2e} 
\usepackage{algorithm, algorithmic}
\usepackage{booktabs}
\usepackage{arydshln}
\usepackage{amsmath}
\usepackage{subcaption}
\usepackage{listings}
\usepackage{spverbatim}
\usepackage{newfloat}
\usepackage{listings}

\usepackage[T1]{fontenc}
\usepackage[utf8]{inputenc}

\usepackage{microtype}

\usepackage{inconsolata}

\usepackage{fdsymbol}
\newcommand\blfootnote[1]{%
  \begingroup
  \renewcommand\thefootnote{}\footnote{#1}%
  \addtocounter{footnote}{-1}%
  \endgroup
}
\title{ASPIRE: Language-Guided Data Augmentation for Improving Robustness Against Spurious Correlations}

\author{
    Sreyan Ghosh\textsuperscript{\rm*},
    Chandra Kiran Reddy Evuru\textsuperscript{\rm*},
    Sonal Kumar,
    Utkarsh Tyagi,
    S Sakshi,\\
    \bf Sanjoy Chowdhury,
    \bf Dinesh Manocha \\
    \textsuperscript{\rm 1}University of Maryland, College Park, USA \\
    \texttt{\{sreyang,ckevuru,sonalkum,utkarsht,fsakshi,sanjoyc,dmanocha\}@umd.edu} \\
}

\begin{document}
\maketitle
\begin{abstract}
Neural image classifiers can often learn to make predictions by overly relying on non-predictive features that are spuriously correlated with the class labels in the training data. This leads to poor performance in real-world atypical scenarios where such features are absent. This paper presents \textbf{ASPIRE} (Language-guided Data \textbf{A}ugmentation for \textbf{SP}ur\textbf{I}ous correlation \textbf{RE}moval), a simple yet effective solution for supplementing the training dataset with images \textit{without} spurious features, for robust learning against spurious correlations via better generalization. ASPIRE, guided by language at various steps, can generate non-spurious images without requiring any group labeling or existing non-spurious images in the training set. Precisely, we employ LLMs to first extract foreground and background features from textual descriptions of an image, followed by advanced language-guided image editing to discover the features that are spuriously correlated with the class label. Finally, we personalize a text-to-image generation model using the edited images to generate diverse in-domain images \textit{without} spurious features. ASPIRE is complementary to all prior robust training methods in literature, and we demonstrate its effectiveness across 4 datasets and 9 baselines and show that ASPIRE improves the worst-group classification accuracy of prior methods by 1\% - 38\%. We also contribute a novel test set for the challenging Hard ImageNet dataset~\footnote{Code and data: https://github.com/Sreyan88/ASPIRE}.

\blfootnote{${^*}$Equal Technical Contribution.} 
\end{abstract}

\section{Introduction}

Spurious correlations are unintended associations or biases learned by models between the input image and the target label, often resulting from factors like data selection biases \cite{torralba2011unbiased,jabri2016revisiting}. The repeated co-occurrence of certain features (like foreground objects or backgrounds), with a more than \textit{average} chance, within instances of a particular class leads the model to learn shortcuts and focus on these spurious non-predictive features for prediction than core ones. For example, most of the images in ImageNet dataset~\cite{deng2009imagenet} labeled as \textit{Dog Sled} also show a dog, and image classifiers trained on ImageNet fail to correctly identify an image of a dog sled \textit{without} a dog in it.

\begin{figure}[t]
    \centering
    \includegraphics[width=\columnwidth]{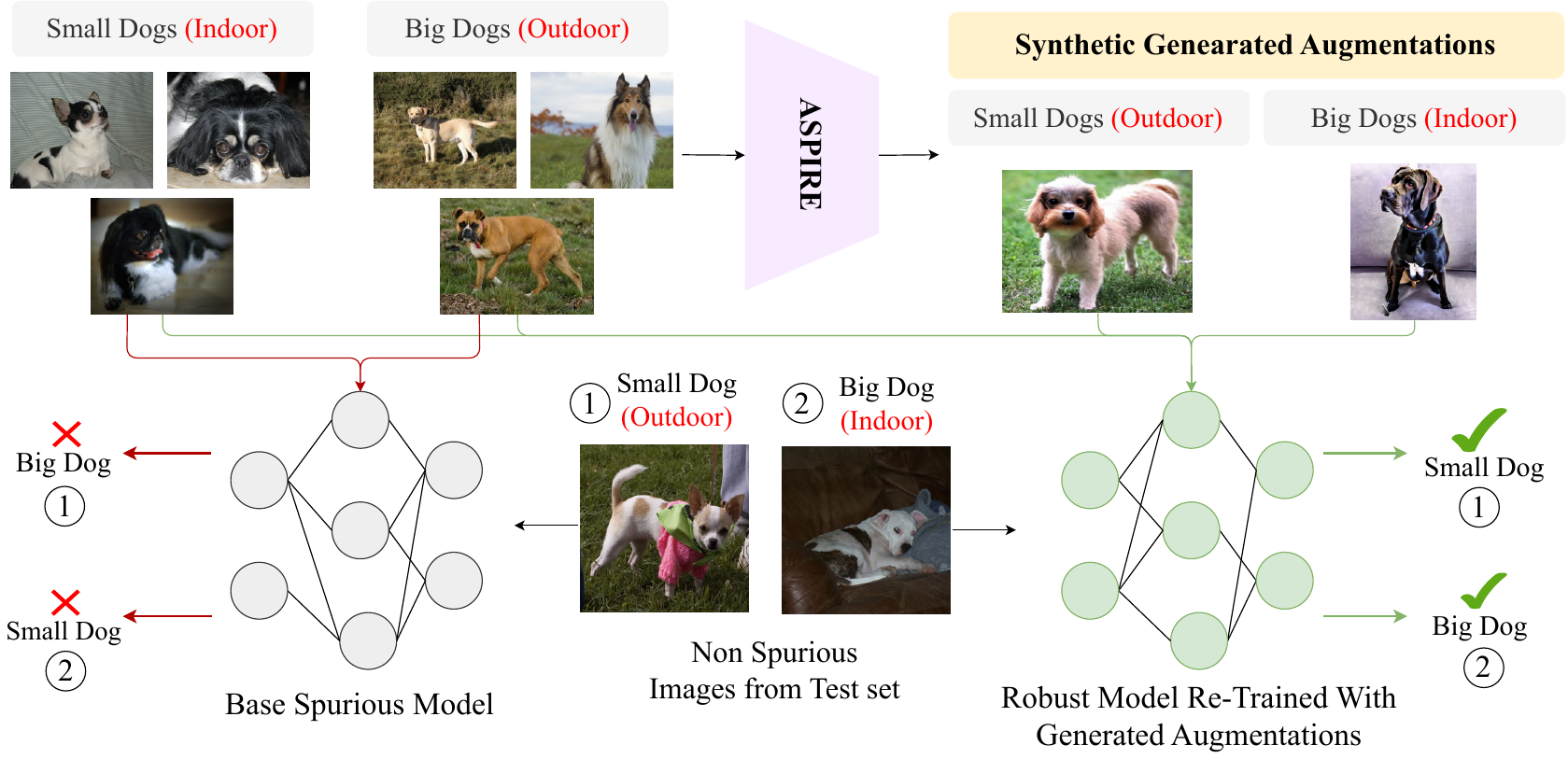}
    \caption{\small Overview of \textbf{ASPIRE}. Given a training dataset, ASPIRE automatically detects non-predictive spuriously correlated features for each class (e.g., indoor background for small dogs) and generates synthetic images without them (small dogs in an outdoor background). These images can then be added to the train set to learn a more robust image classifier.}
    \label{fig:brief}
\end{figure}

Instances of a class in the training set where the co-occurring spurious features are present are commonly known as \textit{majority groups}, while atypical instances where such spurious features are absent are known as \textit{minority groups}. Deep neural networks trained on these datasets poorly generalize on minority groups (naturally due to their scarcity) and thus can exhibit significant performance degradation on minority groups in the test \cite{sagawa2019distributionally}, or in real-world scenarios when encountering domain shift \cite{arjovsky2019invariant}. When training over-parameterized deep neural networks, there are multiple solutions with the same loss values at any given training stage, and the optimizer usually gravitates towards a solution with lesser complexity (or tends to learn a shortcut) \cite{wilson2017marginal,valle2018deep,arpit2017closer,kalimeris2019sgd}. When faced with co-occurring spurious features, the optimizer may preferentially utilize them, as they often require less complexity than the anticipated semantic signals of interest \cite{bruna2013invariant,bruna2015intermittent,brendel2019approximating,khani2021removing}. Even powerful classifiers like CLIP and ViT undergo a significant drop in performance when exposed to minority group images in the test~\cite{yang2023mitigating, kirichenko2023last}. Teaching meaningful data representations to deep neural networks that yield good performance on a target downstream task while avoiding over-reliance on spurious features remains a central challenge in CV.

{\noindent \textbf{Motivation.}} Learning classifiers robust to spurious correlations is an active area of research \cite{Sagawa*2020Distributionally,pmlr-v139-liu21f,kirichenko2023last}, and has the potential to improve various Computer Vision (CV) applications such as visual question-answering \cite{liu2023cross}, retrieval \cite{kong2023mitigating,kim2023exposing}, classification \cite{pmlr-v139-liu21f}, etc. have shown to consistently  In prior work, researchers generally employed different learning techniques with the assumption that annotated data for the minority groups existed in the training dataset. Most of these works are built on the same base principle: improved generalization on minority groups can lead to a more robust classifier. Despite extensive research in deep learning indicating that more data may lead to better generalization, little effort has been made to leverage this principle specifically 
for building robust classifiers. Additionally, we argue that it is impractical to manually collect and label minority group images for real-world, large-scale datasets. For example, in more complex datasets like the Hard ImageNet, beyond the commonly evaluated CelebA \cite{liu2015deep} and Waterbirds \cite{399}, a single class of images may have multiple spuriously correlated features. Thus, identifying all such features through human perception to collect and label minority group images is a difficult task. 
\vspace{0.5mm}

{\noindent \textbf{Main Contributions.}} In this paper, we present ASPIRE, a novel technique to augment existing image classification datasets with diverse non-spurious images for building robust image classifiers. Intuitively, our solution exploits the fact that more data can lead to \textit{better generalization} on minority groups \cite{sagawa2020distributionally,liu2021just}. Guided by language, ASPIRE does not depend on any additional image annotations or human-labeled non-spurious data and only requires a training dataset and a standard model trained using Empirical Risk Minimization (ERM) to identify most of the spurious correlated features for each class in the training dataset. ASPIRE first selects a small portion of the total instances in the training set, misclassified by a classifier after ERM training. These selected images are then captioned, and an LLM extracts the tokens from the caption that describe the foreground objects and background. This is followed by editing the image using advanced language-guided image editing pipelines to remove or replace one object at a time and predicting the class of the edited image using the standard ERM classifier. We attribute the objects or background features that lead to the highest miss-prediction (due to its absence) as \textit{plausible} spurious correlations learned by the model. Finally, we personalize a diffusion model on the edited images to generate diverse in-domain synthetic images for each class with our desired features, i.e., without the \textit{plausible} spurious correlations detected by ASPIRE. To summarize, our main contributions are as follows:

\begin{itemize}
    \item We propose ASPIRE, a method to expand existing datasets with non-spurious images to build more robust image classifiers. ASPIRE is dataset-agnostic (works with any dataset with one or multiple spuriously correlated features per class), training-method agnostic (complements all other methodologies proposed in prior work), and does not need any additional labeled supervision of spurious features or non-spurious images.
    \item  We extensively evaluate ASPIRE on 4 datasets and 9 baselines and show that augmentations generated by ASPIRE improve the worst-group accuracy of all baselines. Additionally, we perform extensive qualitative analysis to prove the effectiveness of ASPIRE.
    \item We contribute a novel test set for the Hard ImageNet dataset \cite{moayeri2022hard} equally balanced with spurious and non-spurious images to promote research in this space.
\end{itemize}

\begin{figure*}[t]
    \centering
    \includegraphics[width=\textwidth]{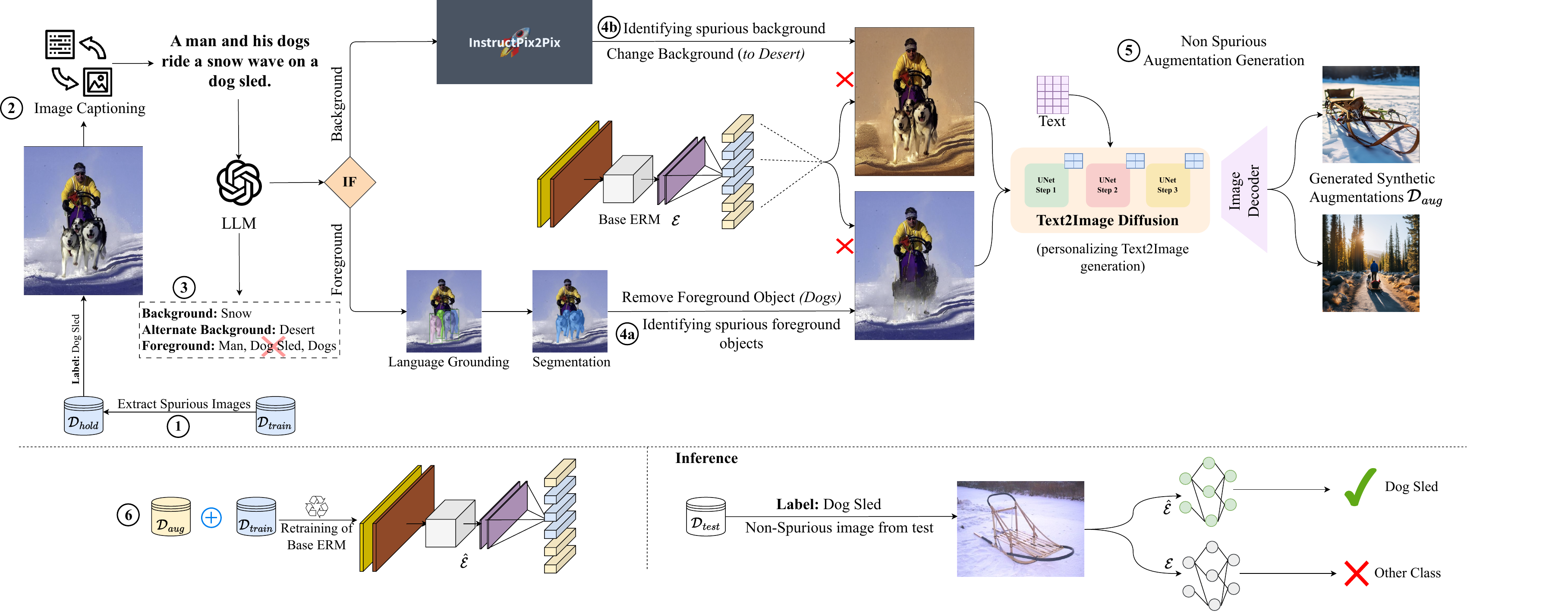}
    \caption{\small Illustration of \textbf{ASPIRE}: ASPIRE follows a 6-step process to improve the robustness against spurious correlations. \Circled{1} We  first train a base classifier $\mathcal{E}$ using ERM on the entire training set and extract images with features that are spuriously correlated to construct $\mathcal{D}_{hold}$. \Circled{2} We caption each image in $\mathcal{D}_{hold}$. \Circled{3} We feed the caption to a LLM and extract the foreground objects and background for each image. \Circled{4a} We remove one foreground object at a time and predict the class of the edited image $\mathcal{E}$. If $\mathcal{E}$ predicts incorrectly, we consider the object as a plausible spurious correlation learned by $\mathcal{E}$ for that class. \Circled{4b} We edit the image to change its original background with an alternative background suggested by the LLM and follow the process to similar to \Circled{4a}. \Circled{5} We personalize a text-to-image diffusion model using edited images from the previous step for the top-\textit{k} unique items leading to the highest number of wrong predictions. \Circled{6} We re-train $\mathcal{E}$ using the generated augmentations to obtain $\hat{\mathcal{E}}$.}
    \label{fig:diagram}
\end{figure*}

\section{Methodology}
\label{sec:method}
\noindent{\textbf{Preliminaries.}} In this section, we provide an overview of our proposed approach. Fig. \ref{fig:diagram} pictorially describes the various steps in ASPIRE. Let's assume we have a training dataset $\mathcal{D}_{train}$ = \{$x_i$,$y_i$\}, where every group of images belonging to a particular class predominantly has images with co-occurring spurious features, also known as the majority group. $\mathcal{D}_{train}$ might have a much smaller number of non-spurious images, or might not, which is also known as the minority group. We do not assume our training dataset to have any group labeling or additional supervision, like labeling for spurious objects. We also have a test dataset $\mathcal{D}_{test}$ = \{$x_i$,$y_i$\} where both groups are represented equally. Additionally, we have a model $\mathcal{E}$ trained on $\mathcal{D}_{train}$, using naive Empirical Risk Minimization (ERM). Thus, $\mathcal{E}$ would already identify the majority group images in $\mathcal{D}_{test}$ with remarkable accuracy; our primary objective is to improve the performance of the classifier on the minority group without hurting the model's overall performance. The next subsections describe each step in detail.
\vspace{0.5mm}

\noindent{\textbf{(1) Extracting $\mathcal{D}_{hold}$ using $\mathcal{E}$.} We use $\mathcal{E}$ to extract a small hold-out set from $\mathcal{D}_{train}$, which we denote as $\mathcal{D}_{hold}$. $\mathcal{D}_{hold}$ should consist of images with spurious correlations in the train (or the majority group), which we will use in our later stages to detect the specific features that are the spurious correlations. Precisely, we first identify training examples that are correctly classified by a standard ERM model and then randomly select p\% of them for constructing $\mathcal{D}_{hold}$. We are inspired by prior work in this space and build on the heuristic that a well-trained classifier tends to have low majority group loss (and subsequently high majority group accuracy) \cite{pmlr-v139-liu21f,nam2020learning}.
\vspace{0.5mm}

\noindent{\textbf{(2) Image Captioning on $\mathcal{D}_{hold}$.} As mentioned earlier, ASPIRE depends on language guidance to achieve its primary objective of generating synthetic, non-spurious images. Thus, in this step, we generate a textual description of each image in $\mathcal{D}_{hold}$, which can capture foreground and background information in the image. To accomplish this, we use a state-of-the-art image captioning model, GIT~\cite{wang2022git}. We expect our image description to include information about most of the visible foreground objects and the predominant background, and we found captions generated by GIT to meet these requirements and not suffer from spurious correlations themselves. As captioning tools get better, we acknowledge that replacing GIT with its improved counterparts will improve the performance of ASPIRE even further.
\vspace{0.5mm}

\noindent{\textbf{(3) Extracting objects and backgrounds from captions.} After captioning, we use LLMs to extract the phrases in the caption that correspond to foreground and background objects and the single predominant background. We assume our search space for identifying spurious correlations to be bounded within them, which is a reasonable assumption in most real-world cases and also in line with most prior work in this space \cite{joshi2023towards}. Recent LLMs have been shown to possess superior reasoning abilities, and we employ GPT-4 for our task~\cite{openai2023gpt4}. LLaMa-2 70B~\cite{touvron2023llama} also proved to be competitive in this task. However, GPT-4 made fewer mistakes. An example of the input and output of this step of ASPIRE is as follows: \textit{Original Caption}: ``A man with two dogs and a sled in the snow.", \textit{Original Label}: ``Dog Sled". \textit{Output:} {foreground: [``man",``dogs"], background:[``snow"], alternate background:[``desert"]}. For simplicity, let us denote the list of identified foreground and background objects for image $x_i$ as $\mathcal{F}_i$ and the predominant background as $\mathcal{B}_i$ (more about the alternate background in Step \textbf{4.b.}). The task of extracting objects and backgrounds from text captions is effectively an information extraction task that involves understanding the structure of the sentence and the relationship between the words, and we found LLMs to deal better with anomalies and out-of-distribution text scenarios than traditional NLP methodologies (algorithmic details about the traditional NLP method can also be found in Appendix \ref{sec:algorithm}). Additionally, we want the identified objects or background to ignore the actual class label. This is crucial for the ASPIRE pipeline, as we do not want to edit the core feature in the image (discussed in detail in the next subsections). This can also be challenging as the class label may or may not exactly appear in the caption. However, we found LLMs to complete this task with remarkable accuracy and ASPIRE to be able to handle minor errors (due to top-\textit{k} selection described later in this section). We use a single generic prompt with annotated exemplars for all datasets, which can be found in Appendix \ref{subsec:llama}.
\vspace{0.5mm}

\noindent{\textbf{(4.a.) Identifying spurious foreground objects.} The primary objective of this step is to identify one or several unique features per class that are plausible spurious correlations. To achieve this, we build on recent advancements in language-guided image in-painting to remove one object at a time identified in $\mathcal{F}_i$ followed by allowing $\mathcal{E}$ to predict the class of the edited image. If $\mathcal{E}$ predicts the image correctly, we do not do anything with that image. If $\mathcal{E}$ predicts an image incorrectly, we identify the removed object as a plausible spurious correlation for the class $c$ in the dataset and add the image to a set $\mathcal{D}_{synth}$ (which we later use to personalize text-to-image generation). Additionally, we add the text phrase of the spurious object to another set $\mathcal{T}_{synth}$.

Precisely, for every fore-ground object $f$ in $\mathcal{F}_i$, we first localize the object using Grounding DINO \cite{liu2023grounding}, which takes as input the text phrase of $f$ identified from the caption and outputs a bounding box $bb$ for $f$. This is followed by extracting the segmentation map $\mathcal{M}$ for $f$ using Segment Anything \cite{kirillov2023segment}, which accepts $bb$ as the segmentation prompt. $\mathcal{M}$ is then used to remove $f$ from $x_i$ using LaMa image in-painting \cite{suvorov2022resolution}. For detailed information on the workings of Grounding DINO, Segment Anything, and LaMa, we request our readers refer to the original paper.
\vspace{0.5mm}

\noindent{\textbf{(4.b.) Identifying spurious backgrounds.} The primary objective of this step is to identify if the predominant background of the image $x_i$ serves as a spurious correlation for the particular class $c$ of images in the dataset to which $x_i$ belongs. Following a hypothesis similar to (\textbf{4.a.}), we assume that if removing the background $b$ in $\mathcal{B}_i$ from $x_i$ can lead $\mathcal{E}$ to a wrong prediction, $b$ can be a plausible spurious correlation. However, removing the background altogether (and keeping just the foreground items) not only disrupts the image semantics but is also not representative of real-world cases. Prior work also shows that removing the background forces the model to pay attention to foreground objects \cite{kirichenko2023last} which are not suitable for our use case. Thus, we employ recent advances in instruction-based image editing to achieve this task.  We first leverage the superior reasoning abilities of an LLM to suggest an alternate contrasting background $\tilde{b}$ for the image from its caption. Next, we instruct InstructPix2Pix \cite{brooks2023instructpix2pix} to convert the background of $x_i$ from $b$ to $\tilde{b}$. Similar to the previous step, if $\mathcal{E}$ predicts the image correctly, we do not do anything with that image. However, if $\mathcal{E}$ predicts an image incorrectly, we identify the original background as a plausible spurious correlation and add the edited image to $\mathcal{D}_{synth}$ while we add the original text phrase of the background from the caption to $\mathcal{T}_{synth}$.
% captions, we ask an LLM once again to identify a 
\vspace{0.5mm}

\noindent{\textbf{(5) Non-spurious augmentation generation.} The primary objective of this step is to generate in-domain non-spurious images $\mathcal{D}_{aug}$ for every class in the dataset $\mathcal{D}_{train}$. These generated augmentations can then be used to supplement the training dataset $\mathcal{D}_{train}$ followed by re-training $\mathcal{E}$ to reduce its reliance on the spurious correlations. Generating \textit{in-domain} augmentations without non-spurious features is crucial to the success of our approach as out-of-distribution samples may adversely affect model performance \cite{trabucco2023effective}. The most trivial approach would be to generate $\mathcal{D}_{aug}$ by prompting any open-source text-to-image model. However, there exist two primary roadblocks to this approach: \textbf{(1)} Open-source diffusion models trained on internet-scale data generate diverse images for diverse prompts. Thus, prompting these models does not confirm the consistency of generations with the underlying distribution. \textbf{(2)} These models also posses spurious correlations or biases themselves \cite{trabucco2023effective}. For example, prompting Stable Diffusion with the prompt: \textit{``picture of a dog sled''} generates dog sleds with dogs most of the time. Attaching negative words with the prompts (like \textit{``picture of a dog sled without a dog''}) often leads to the same spurious images as Stable Diffusion (and most image generation models) are known to not adhere to such negation in prompts~\cite{tong2024mass}.

To overcome the aforementioned problems and generate in-domain images with the desired non-spurious features, we resort to personalizing a text-to-image generation model. Specifically, we train Stable Diffusion using textual-inversion \cite{gal2023an} with samples from top-\textit{k} phrases in $\mathcal{T}_{synth}$, and their corresponding images in $\mathcal{D}_{synth}$. Textual-inversion effectively learns concepts and style from a small set of images for each class in $\mathcal{D}_{synth}$ by just fine-tuning a single token in the embedding layer (which in our case is just the original class label) without over-fitting the generation model. $\mathcal{D}_{synth}$ is the perfect candidate for extracting this small set as it contains non-spurious images, i.e., images without spurious features and concepts. Finally, we prompt the fine-tuned model to generate $n\times$ diverse samples for $\mathcal{D}_{aug}$.
\vspace{0.5mm}

{\noindent \textbf{Top-\textit{k} selection.}} Recall that $\mathcal{D}_{synth}$ and their corresponding text phrases in $\mathcal{T}_{synth}$ represent \textit{all} wrongly predicted edited instances, i.e., they have a diverse set of foreground objects and backgrounds for each class. Thus, we attribute only the top-\textit{k} unique items in $\mathcal{T}_{synth}$ with the highest frequencies as the spurious correlation associated with that class and use images from only the top-\textit{k} items for diffusion personalization. However, due to diversity in generated captions, text phrases corresponding to the same type of objects and backgrounds may be represented in $\mathcal{T}_{synth}$ in diverse forms, for e.g., [``dogs'',``dog'',``two dogs'', $\cdots$]. Thus, before selecting the top-\textit{k} items, we first collapse all the similar phrases to one by first finding the root for all phrases in $\mathcal{T}_{synth}$ by stemming and then calculating the cosine similarity between the glove embedding of the roots (to account for dissimilar roots, for e.g., ``snow'' and ``snowy mountain''). Items with a cosine similarity of $\geq$ 0.90 are collapsed into one.
\vspace{0.5mm}

{\noindent \textbf{(6) Re-training the base classifier $\mathcal{E}$.}} Once we have generated $\mathcal{D}_{aug}$, we add the generated images to the existing $\mathcal{D}_{train}$ to re-train our standard classifier $\mathcal{E}$. As mentioned earlier, the ASPIRE augmentation methodology is training-method-agnostic, and the augmentations generated can be coupled with any existing training approach from literature. The next Section describes how we add ASPIRE augmentations to our baseline training pipelines.

\begin{table*}[t!]
    \center
    \resizebox{1\textwidth}{!}{\renewcommand{\arraystretch}{1.2}
\begin{tabular}{lccccccccc}
% \label{tab:main}
\toprule

\multicolumn{1}{c}{Method} &
  \multicolumn{2}{c}{Waterbirds} &
  \multicolumn{2}{c}{CelebA} &
  \multicolumn{2}{c}{SpucoDogs} &
  \multicolumn{2}{c}{Hard ImageNet} \\ 
  \cmidrule(lr){2-3} \cmidrule(lr){4-5} \cmidrule(lr){6-7} \cmidrule(lr){8-9}
  &
  Worst-group Acc. (\%) &
  Avg Acc. (\%) &
  Worst-group Acc. (\%) &
  Avg Acc. (\%) &
  Worst-group Acc. (\%) &
  Avg Acc. (\%) &
  Worst-group Acc. (\%) &
  Avg Acc. (\%) \\ \midrule
ERM        & 74.4                                          & 96.9 & 43.4                                          & 95.5 					& 42.3                                          & 74.5                           & 12.6                                           & 74.3                           \\
ERM + \citeauthor{azizi2023synthetic}    & 71.8 & \textbf{97.1}   &  36.2					&  \textbf{96.7}                                        & 39.6                           &  75.4                                         &  10.7 & 76.7                          \\
ERM + \citeauthor{gowal2021improving}  & 75.7 & 85.6   &  45.7					&  96.4                                        & 46.8                          &  73.7                                         &  23.3 & 83.4                          \\ \hdashline
ERM + ASPIRE  & \textbf{$\textbf{78.7}_{\pm1.31}$ \textcolor{ForestGreen}{(+4.3)}} & $89.6_{\pm1.10}$          & \textbf{$\textbf{50.5}_{\pm0.79}$ \textcolor{ForestGreen}{(+7.1)}} & $95.4_{\pm1.08}$                           & \textbf{$\textbf{51.6}_{\pm0.86}$ \textcolor{ForestGreen}{(+9.3)}} & \textbf{$\textbf{75.5}_{\pm1.18}$} 				 & \textbf{$\textbf{50.1}_{\pm1.26}$ \textcolor{ForestGreen}{(+37.5)}} & \textbf{$\textbf{96.5}_{\pm1.32}$} 				   \\\hline
LfF \citep{nam2020learning}  & 78.0                                          & 91.2          & 77.2                                          & 85.1                           & 70.2                                          & 80.8                           & 58.8                          & 92.5                           \\
LfF + \citeauthor{azizi2023synthetic}   & 74.2 & \textbf{92.3}   & 74.4 &  85.7   &   67.5  &  \textbf{81.6} & 54.3  & 92.6 \\
LfF + \citeauthor{gowal2021improving}  &  81.0    &  89.3    & 78.2 & 85.8  &  72.9 & 80.9 & 60.3    &  92.7    \\ \hdashline
LfF + ASPIRE  & \textbf{$\textbf{83.2}_{\pm0.20}$ \textcolor{ForestGreen}{(+5.2)}} & \textbf{$91.4_{\pm1.12}$} & \textbf{$\textbf{81.7}_{\pm0.43}$ \textcolor{ForestGreen}{(+4.5)}} & \textbf{$\textbf{86.3}_{\pm1.25}$} 					& \textbf{$\textbf{75.4}_{\pm0.38}$ \textcolor{ForestGreen}{(+5.2)}} & \textbf{$80.9_{\pm0.31}$}                           & \textbf{$\textbf{63.8}_{\pm0.30}$ \textcolor{ForestGreen}{(+5.0)}}  & \textbf{$\textbf{92.7}_{\pm0.21}$}                           \\
\hline
Group DRO \cite{sagawa2019distributionally}  & 91.4                                          & 93.5          & 88.9                                          & 92.9                           & 75.4                                          & 82.8                           & 65.6                                           & 91.8                           \\
Group DRO + \citeauthor{azizi2023synthetic}    &  88.2 &    94.1&  85.6&      93.2&                            71.7&                                          84.1&   62.8&                           \textbf{92.9}\\
Group DRO + \citeauthor{gowal2021improving}  &      91.6&      94.2&   					89.8&                                          93.7&                           76.3&                                           83.4&     65.5&                           91.7\\ \hdashline
Group DRO + ASPIRE    & \textbf{$\textbf{92.8}_{\pm0.49}$ \textcolor{ForestGreen}{(+1.4)}} & \textbf{$\textbf{94.6}_{\pm0.49}$}          & \textbf{$\textbf{90.1}_{\pm1.08}$ \textcolor{ForestGreen}{(+1.2)}} & \textbf{$\textbf{94.3}_{\pm0.92}$}                           & \textbf{$\textbf{78.7}_{\pm1.26}$ \textcolor{ForestGreen}{(+3.3)}} & \textbf{$\textbf{84.3}_{\pm0.58}$}                           & \textbf{$\textbf{67.4}_{\pm1.01}$ \textcolor{ForestGreen}{(+1.8)}}  & \textbf{$92.4_{\pm0.59}$}                           \\ \hline
JTT \citep{liu2021just}  & 86.7                                          & 93.3          & 81.1                                          & 88.0                           & 73.0                                          & 80.4                           & 63.5                                           & 90.6                           \\
JTT + \citeauthor{azizi2023synthetic}   &  83.2&    \textbf{94.9} &  					78.3&                                          90.2&                            71.8&                                          \textbf{82.2}&   61.4&                           92.4\\
JTT + \citeauthor{gowal2021improving}     & 87.5 &      94.2&   					83.8&                                          89.6&                           74.1&                                           81.1&     64.1&        91.9\\ \hdashline
JTT + ASPIRE    & \textbf{$\textbf{90.2}_{\pm1.16}$ \textcolor{ForestGreen}{(+3.5)}} & \textbf{$94.6_{\pm1.24}$}          & \textbf{$\textbf{85.7}_{\pm0.64}$ \textcolor{ForestGreen}{(+4.6)}} & \textbf{$\textbf{91.6}_{\pm0.75}$}                          & \textbf{$\textbf{75.5}_{\pm1.33}$ \textcolor{ForestGreen}{(+2.5)}} & \textbf{$81.7_{\pm1.12}$}                           & \textbf{$\textbf{65.2}_{\pm0.54}$ \textcolor{ForestGreen}{(+1.7)}}  & \textbf{$\textbf{92.9}_{\pm0.82}$}                           \\ \hline
DivDis \citep{DBLP:journals/corr/abs-2202-03418}   & 85.6                                          & 87.3          & 55.0                                          & 90.8                           & 39.3                                          & 65.5                           & 15.5                                           & 71.8                           \\
DivDis + \citeauthor{azizi2023synthetic}   &  84.2&    \textbf{88.6} &  					53.7&                                          \textbf{92.2} &                            37.5&                                          66.4&   13.7&                           77.2\\
DivDis + \citeauthor{gowal2021improving}    &      86.3 &      87.4&   					56.1&                                          91.2&                           42.1&                                           66.3&     23.9&                           76.9\\ \hdashline
DivDis + ASPIRE   & \textbf{$\textbf{87.2}_{\pm0.49}$ \textcolor{ForestGreen}{(+1.6)}} & $87.8_{\pm0.84}$          & \textbf{$\textbf{57.4}_{\pm1.13}$ \textcolor{ForestGreen}{(+2.4)}}                       & \textbf{$91.6_{\pm0.66}$}                           & \textbf{$\textbf{43.6}_{\pm1.48}$ \textcolor{ForestGreen}{(+4.3)}} & \textbf{$\textbf{67.1}_{\pm1.22}$}                           & \textbf{$\textbf{35.5}_{\pm0.82}$ \textcolor{ForestGreen}{(+20.0)}} & \textbf{$\textbf{77.6}_{\pm0.34}$}                           \\ \hline
SUBG \citep{idrissi2022simple}   & 88.9                                          & 91.2          & 86.2                                          & 89.1                           & 74.2                                          & 81.5                           & 62.3                                           & 90.9                           \\
SUBG + \citeauthor{azizi2023synthetic} &  86.5&    91.8&  					85.4&                                         \textbf{91.3}&                            72.3&                                          81.6&   60.5&                           \textbf{92.9} \\
SUBG + \citeauthor{gowal2021improving}    &      89.7&      91.7&   					88.2&                                          89.9&                           75.6&                                           81.7&     64.8&                           91.6\\ \hdashline
SUBG + ASPIRE   & \textbf{$\textbf{90.7}_{\pm0.62}$ \textcolor{ForestGreen}{(+1.8)}} & \textbf{$\textbf{92.1}_{\pm0.88}$}          & \textbf{$\textbf{88.6}_{\pm1.37}$ \textcolor{ForestGreen}{(+2.4)}} & \textbf{$90.1_{\pm0.64}$}                           & \textbf{$\textbf{77.5}_{\pm0.73}$ \textcolor{ForestGreen}{(+3.3)}} & \textbf{$\textbf{83.5}_{\pm0.92}$}                           & \textbf{$\textbf{66.7}_{\pm1.22}$ \textcolor{ForestGreen}{(+4.4)}}  & \textbf{$92.4_{\pm0.63}$}                           \\ \hline
Correct-n-Contrast \citep{zhang2022correct}   & 88.7                                          & 90.6          & 88.1                                          & 89.4                           & 73.7                                          & 81.2                           & 60.5                                           & 91.7                           \\
Correct-n-Contrast + \citeauthor{azizi2023synthetic}   &  84.3 &    \textbf{93.4} &  					85.2 &  91.3 &  70.8  & \textbf{85.6}   &58.7    &\textbf{93.3}\\
Correct-n-Contrast + \citeauthor{gowal2021improving}   &      89.1  &      91.7&   					88.7  &90.6   &74.9       &82.6       &63.2       &92.1   \\ \hdashline
Correct-n-Contrast + ASPIRE   & \textbf{$\textbf{90.8}_{\pm1.18}$ \textcolor{ForestGreen}{(+2.1)}} & \textbf{$92.6_{\pm1.48}$}          & \textbf{$\textbf{89.9}_{\pm1.45}$ \textcolor{ForestGreen}{(+1.8)}} & \textbf{$\textbf{91.3}_{\pm0.28}$}                           & \textbf{$\textbf{76.8}_{\pm1.10}$ \textcolor{ForestGreen}{(+3.1)}} & \textbf{$83.1_{\pm1.04}$}                           & \textbf{$\textbf{65.9}_{\pm0.94}$ \textcolor{ForestGreen}{(+5.4)}}  & \textbf{$91.9_{\pm1.11}$}                           \\ \hline
MaskTune \citep{taghanaki2022masktune}   & 78.0                                          & 91.2          & 77.9                                          & 92.5                           & 31.6                                          & 59.2                           & 33.0                                          & 58.5                          \\
MaskTune + \citeauthor{azizi2023synthetic}   &  75.8&    \textbf{93.4}   & 73.3  & \textbf{93.5}  & 26.3  & \textbf{63.4}  &   28.9    & \textbf{61.3}\\
MaskTune + \citeauthor{gowal2021improving}     &79.3  &  85.2&    78.8 &                                          88.1  &35.2       &60.7       &     35.3  & 55.8 \\ \hdashline
MaskTune + ASPIRE    & \textbf{$\textbf{81.6}_{\pm1.28}$ \textcolor{ForestGreen}{(+3.6)}} & \textbf{$91.3_{\pm0.54}$}          & \textbf{$\textbf{81.2}_{\pm0.22}$ \textcolor{ForestGreen}{(+3.3)}} & \textbf{$92.8_{\pm0.38}$}                           & \textbf{$\textbf{37.5}_{\pm0.33}$ \textcolor{ForestGreen}{(+5.9)}} & \textbf{$61.3_{\pm1.05}$}                           & \textbf{$\textbf{41.0}_{\pm0.61}$ \textcolor{ForestGreen}{(+8.0)}} & \textbf{$60.2_{\pm0.37}$}                           \\ \hline
DFR \citep{kirichenko2023last}  & 81.7                                          & 90.1          & 80.5                                          & 85.3                           & 78.8                                          & 83.2                           & 33.3                                           & 95.7                           \\
DFR + \citeauthor{azizi2023synthetic}   &  78.6 &    \textbf{92.7}&  					78.3&                                          88.4&                            72.1&                                          85.1 &   29.5&                           \textbf{96.3}\\
DFR  + \citeauthor{gowal2021improving} &      83.1&      86.5&   					83.4&                                          86.2&                           81.0&                                           84.4&     35.2 &   92.0                        \\ \hdashline
DFR + ASPIRE   & \textbf{$\textbf{85.3}_{\pm1.34}$ \textcolor{ForestGreen}{(+3.6)}} & \textbf{$91.7_{\pm0.79}$}          & \textbf{$\textbf{85.5}_{\pm0.64}$ \textcolor{ForestGreen}{(+5.0)}} & \textbf{$\textbf{89.5}_{\pm0.51}$}                           & \textbf{$\textbf{84.2}_{\pm0.83}$ \textcolor{ForestGreen}{(+5.4)}}                        & \textbf{$\textbf{87.5}_{\pm0.57}$}                           & \textbf{$\textbf{37.5}_{\pm0.39}$ \textcolor{ForestGreen}{(+4.2)}}  & \textbf{$96.2_{\pm0.91}$}                           \\ \hline
\bottomrule
\end{tabular}
}

\caption{\small Average and worst-group test accuracies of all baselines trained with and without ASPIRE augmentations. ASPIRE substantially improves the worst-group accuracy of all baselines (in the range of 1\% - 38\%) with just 1$\times$ more augmentations.}
\label{tab:main}
\vspace{-0.5em}
\end{table*}
\section{Experimental Setup}
\label{sec:experiments}

{\noindent \textbf{Datasets.}} To evaluate the effectiveness of ASPIRE, we experiment on 4 benchmark datasets, including Waterbirds \cite{sagawa2019distributionally}, CelebA \cite{liu2015deep}, SPUCO Dogs \cite{joshi2023towards} and Hard ImageNet \cite{moayeri2022hard}. The Waterbirds dataset, generated synthetically by
combining images of birds from the CUB dataset \cite{wah2011caltech} and backgrounds from the Places dataset \cite{zhou2017places}, has 4 groups of images in training and testing datasets including waterbirds on water background, waterbirds on land background, landbirds on water background and landbirds on land background. The minority groups for the dataset (groups with the least number of samples in the training set) are waterbirds on land and landbirds on water. The main challenge is correctly identifying the minority groups in the test. For CelebA, we perform the hair color prediction task, which has 4 groups of images, including blond and non-blond males and blond and non-blond females. The minority group is blond males. SPUCO Dogs has 4 groups of images, including big dogs in indoor and outdoor settings and small dogs in indoor and outdoor settings. The minority groups are big dogs indoors and small dogs outdoors. The Hard ImageNet dataset has images from 15 ImageNet synsets and is more complex than the other 3 datasets, does not have group labeling, and has multiple spurious correlations for each class. For more details, we request our readers to refer to \citet{moayeri2022hard}. Since the dataset does not have a test set, we contribute a novel expert-annotated test dataset with 25 spurious and 25 non-spurious images per class. The spurious and non-spurious features for each class were inspired by the original paper. More details about dataset statistics and annotation can be found in Appendix \ref{subsec:dataset_details}.
\vspace{0.5mm}

{\noindent \textbf{Baselines.}} To prove the efficacy of ASPIRE augmentations, we add ASPIRE augmentations to the original training pipeline for various robust training methods proposed in literature. Precisely, we employ Group DRO~\cite{sagawa2019distributionally}, SUBG~\cite{idrissi2022simple}, Just Train Twice (JTT), Learning from Failure (LfF)~\cite{nam2020learning}, Correct-n-Contrast (CnC) \cite{zhang2022correct}, Deep Feature Reweighting (DFR). \cite{kirichenko2023last} and MaskTune \cite{asgari2022masktune}, To this list, we add the standard Empirical Risk Minimization (ERM) baseline, trained using SGD without any additional modifications. Additionally, we compare ASPIRE augmentations with augmentations generated using the methods proposed by \citet{gowal2021improving} and \citet{azizi2023synthetic}. More details on baselines and how ASPIRE augmentations were added for training can be found in Appendix~\ref{subsec:details}. We do not experiment with LVLMs like LLaVa~\cite{liu2023llava} as there is no simple method to fine-tune them for robustness against spurious correlations proposed in literature. ASPIRE is only meant to complement methods proposed on the standard framework of fine-tuning vision encoders for image classification

\begin{table}[t]
    \centering
    \resizebox{0.75\columnwidth}{!}{
    \begin{tabular}{c|c|c}
    \hline \hline
                      & Worst-group Acc.(\%) & Avg. Acc.(\%) \\ \hline
ASPIRE - Step 4.a.    &   70.65     &   86.54     \\
ASPIRE - Step 4.b.    &   66.40     &   82.67    \\
ASPIRE - Step 5.      &  65.75      &    81.44    \\ \hline
ASPIRE                &  \textbf{71.80}      &  \textbf{87.39}      \\ \hline
    \end{tabular}}
    \caption{\small Ablation study of ASPIRE. ``-'' indicates that the step was removed from the ASPIRE pipeline. All results are averaged across all datasets.}
    \label{tab:ablation}
    \vspace{-1em}
\end{table}

{\noindent \textbf{Hyper-parameters.}} For training the base ERM model, we train the model for 100 epochs with a learning rate of $1e^{-3}$ using the SGD optimizer with a weight decay of $1e^{-4}$. For training all other baselines, we use the original hyper-parameter settings proposed by the authors in their original paper. This includes the seed settings and the number of runs for every model. We use just 1~$\times$ augmentations of non-spurious images. Though this is possible for us as all our current datasets are also annotated with group labels, the number of ASPIRE augmentations to be added can be decided using hyper-parameter search, and we noticed no signs of over-fitting till 3$\times$ augmentations (see Appendix 1). For top-\textit{k}, we resort to \textit{k}=3 post a hyper-parameter search among \textit{k}=\{1,2,3,4,5\}. \textit{k}=3 seemed to capture the most major spurious correlations while ignoring the minor ones. Examples of extracted top-\textit{k} can be found in Figure \ref{fig:examples} and Appendix 1. For prompting InstructPix2Pix, we use Text CFG=7.5 and Image CFG=1.5. Prompt in Appendix \ref{subsec:llama}.

\section{Results and Analysis}
\label{sec:results}

\subsection{Quantitative Analysis}
\label{sec:quant}
Table \ref{tab:main} compares the results of 9 baselines trained with and without ASPIRE augmentations. Worst-group accuracy corresponds to the accuracy of minority groups (or non-spurious images) in the test. As we clearly see, with just 1$\times$ augmentations, ASPIRE improves the average accuracy of our baselines by 0.1\%-22.2\% and the worst-group accuracy of our baselines by 1.2\%-37.5\%. \textit{ASPIRE consistently achieves higher gains in worst-group accuracy and only undergoes a slight drop in average accuracy in some settings, which is in line with prior art and our primary motivation of improving robustness against spurious correlations.} We notice the highest gains in Hard ImageNet, a fundamentally more difficult dataset with no minority group images in the training dataset and multiple spurious correlations per class. Our standard ERM model also witnesses the highest gains among all other baselines. On average, our 2-stage training baselines improve by a higher margin on average than 1-stage baselines due to improved explicit generalization over ASPIRE augmentations. The method proposed by \citet{gowal2021improving} consistently underperforms ASPIRE, thereby highlighting that explicitly removing spurious features in the generated dataset improves robustness. On the other hand, the method proposed by ~\citet{azizi2023synthetic} significantly underperforms ASPIRE in worst group accuracy but outperforms in average accuracy in some settings. This is due to the fact that standard data augmentation amplifies spurious correlations already present in the training set as it generates images with similar features to those on which it is conditioned.
\vspace{0.5mm}
\begin{figure}[h!]
    \centering
    \begin{subfigure}{0.22\textwidth}
        \includegraphics[width=\linewidth]{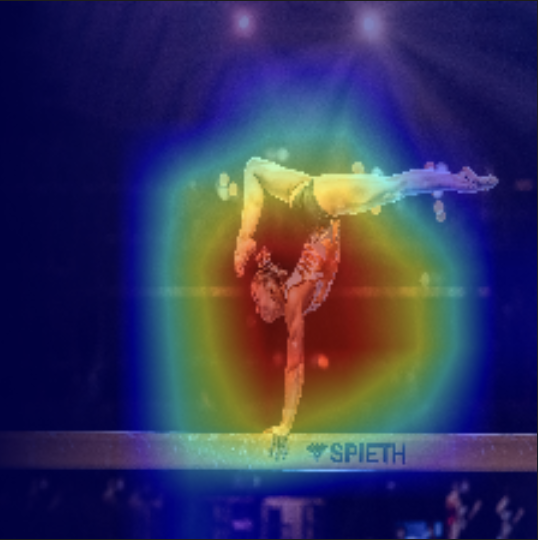}
        \caption{Without augmentations.}
        \label{fig:image1}
    \end{subfigure}
    \hfill
    \begin{subfigure}{0.22\textwidth}
        \includegraphics[width=\linewidth]{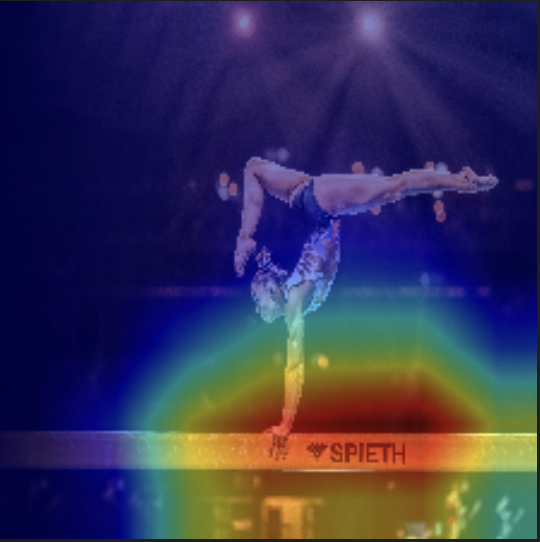}
        \caption{With augmentations.}
        \label{fig:image2}
    \end{subfigure}

    \begin{subfigure}{0.22\textwidth}
        \includegraphics[width=\linewidth]{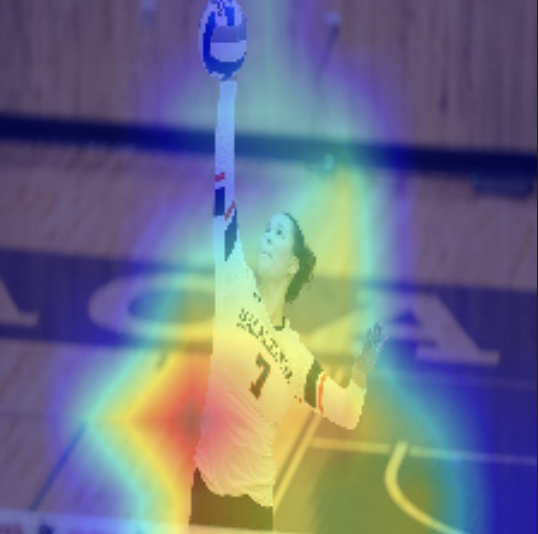}
        \caption{Without augmentations.}
        \label{fig:image3}
    \end{subfigure}
    \hfill
    \begin{subfigure}{0.22\textwidth}
        \includegraphics[width=\linewidth]{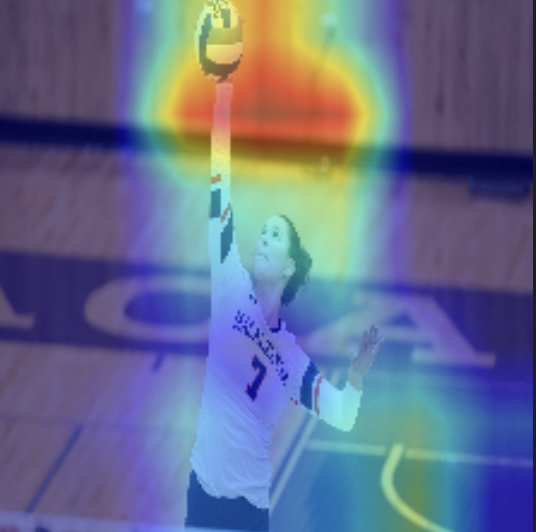}
        \caption{With augmentations.}
        \label{fig:image4}
    \end{subfigure}

    \caption{\small GradCAM visualizations of the features used by the standard ERM model trained with and w/o ASPIRE augmentations on the Hard ImageNet dataset (\textit{Balance beam} top and \textit{Volleyball} bottom). As clearly visible, when trained with ASPIRE augmentations, the model tends to focus better on core features than spurious ones (more in Appendix \ref{sec:gradcam}).}
    % \vspace{-1em}
    \label{fig:four_images}
\end{figure}

{\noindent \textbf{Ablations.}} Table~\ref{tab:ablation} removes certain key components in the ASPIRE pipeline to prove their efficacy. As we can see, the ASPIRE performance decreases significantly when the image generation step is removed (only edited images are used to train the robust classifier). Additionally, ASPIRE undergoes a sharper drop in performance when foreground identification is removed than the background, which we attribute to the design of the test set minority groups of existing datasets.
\begin{figure*}[t!]
\includegraphics[width=\textwidth]{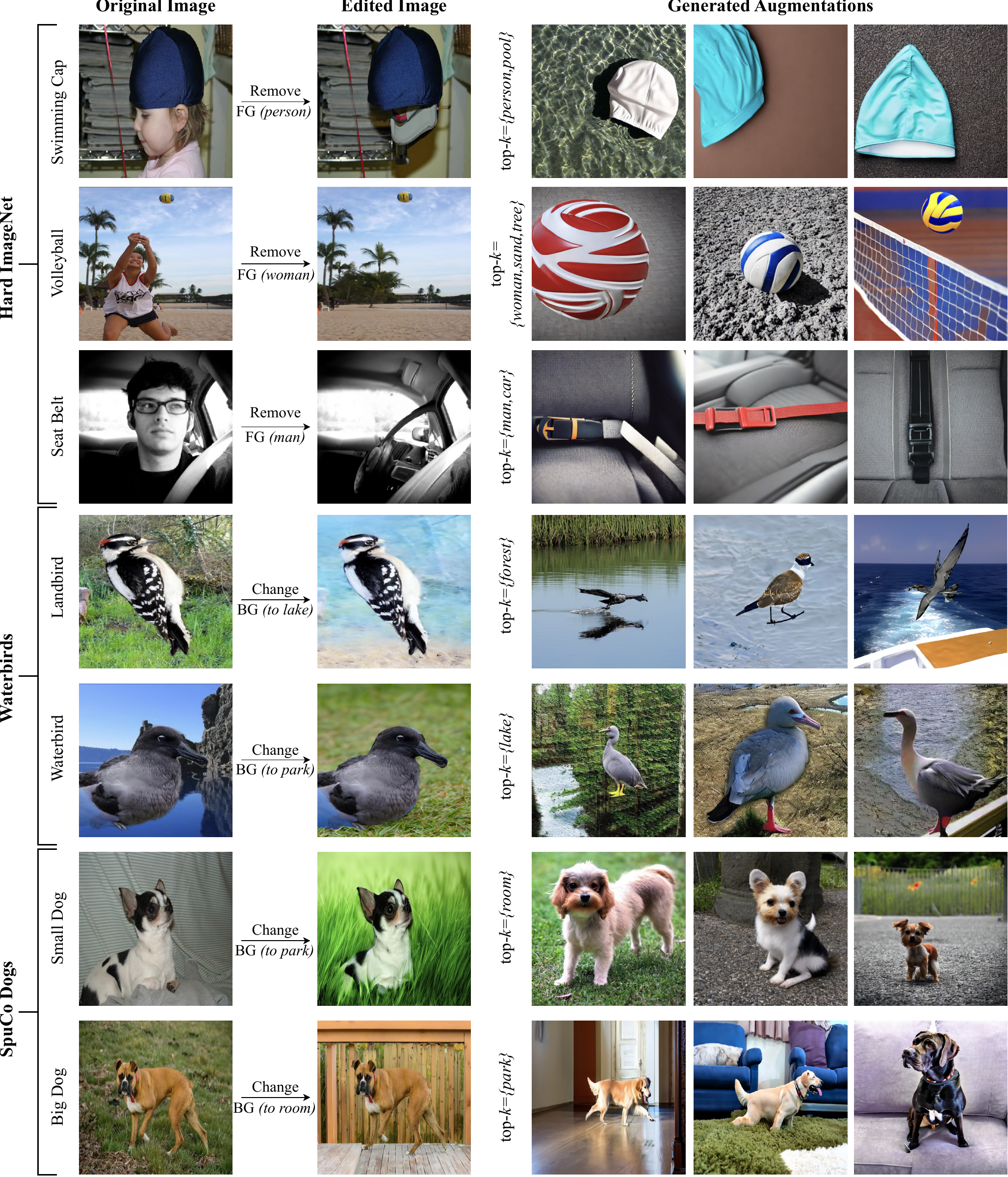}
    \caption{\small Examples of \textbf{Original Images}, \textbf{Edited Images} from the ASPIRE pipeline and \textbf{Generated Augmentations}. To the left of the \textbf{Generated Augmentations}, we also mention the top-\textit{k} spurious correlations discovered by ASPIRE for the particular class. ASPIRE generates diverse augmentations with the desired non-spurious features that can be used to train robust models.}
    \label{fig:examples}
    % \vspace{-2mm}
\end{figure*}

\subsection{Qualitative Analysis}
\label{sec:quality}
Fig. \ref{fig:four_images} illustrates the GradCAM visualizations of the features used by the standard ERM model trained with and w/o ASPIRE augmentations for two classes from the Hard ImageNet dataset, \textit{Volleyball} and \textit{Horizontal Bar}. When trained with ASPIRE augmentations, the model tends to focus better on core features corresponding to the actual class than spurious ones. Fig. \ref{fig:examples} illustrates examples of original images, edited images (edited by the ASPIRE pipeline), and ASPIRE-generated augmentations. ASPIRE successfully captures the major spurious cues learned by a model (shown in top-\textit{k}) and generates diverse images \textit{without} them. We show more examples in Appendix \ref{sec:gradcam} and \ref{sec:generation} and illustrate some failure cases in Appendix \ref{sec:failure}.

\section{Literature Review}
\label{sec:literature}

\citet{Geirhos_2020} provides a detailed survey on how image classification models perform poorly when trained on datasets with spurious correlations. Following this, a lot of works explore SGD training dynamics and inductive biases of such models in the presence of spurious correlations \cite{nagarajan2021understanding,pezeshki2021gradient,rahaman2019on}. \citet{shah2020pitfalls} shows how deep neural networks, trained using ERM, can take shortcuts and learn to rely on spurious features rather than core features for a class. They call this phenomenon the \textit{extreme simplicity bias}. \citet{10.5555/3495724.3496562} and \citet{jacobsen2018excessive} further present examples with both natural and synthetic images, highlighting instances where these networks overlook core features. \citet{shinoda2023shortcut} explore the types of shortcuts that are more likely to be learned.

A plethora of methods in literature propose novel training strategies for improving robustness against spurious correlations \cite{BenTal2011RobustSO,HuNSS18,Sagawa*2020Distributionally,oren-etal-2019-distributionally,zhang2021coping}. A detailed explanation of all these methods can be found in Section~\ref{sec:experiments} and Appendix~\ref{subsec:details}.

The use of synthetic data to improve the performance of downstream CV tasks has been explored extensively in the past. For data-driven generative models, GANs have remained the predominant approach to date \cite{brock2018large,li2022bigdatasetgan}. Very recently, He \textit{et al.}~\cite{he2022synthetic} employ large-scale text-to-image models like GLIDE \cite{nichol2021glide} to augment training data with synthetic images and show improvement in image classification performance. 

Prior work explores language guidance for image generation for varied objectives. For example, \citet{prabhu2023lance} proposes to generate counterfactual images for stress-testing image classification models. On similar lines, \citet{wiles2022discovering} and \citet{vendrow2023dataset} propose to identify failure cases and spurious correlations using augmented data generated using language guidance. Finally, \citet{dunlap2023using} proposes to adapt a model to new domains using augmented data. To the best of our knowledge, generative data augmentation with or without language guidance for improving robustness against spurious correlations has not yet been explored.

\section{Conclusion}
\label{sec:conclusion}
In this paper, we present ASPIRE, a novel data augmentation methodology to augment existing datasets with non-spurious minority group images to build robust and de-biased image classifiers. We evaluate ASPIRE on 4 benchmark datasets with 9 baselines and show that ASPIRE augmentations improve the worst-group accuracy of all baselines while maintaining average accuracies.

\section*{Limitations and Future Work}
\label{sec:limitations}

As part of future work, we would like to address the current limitations of ASPIRE, which include: 
\begin{enumerate}
    \item ASPIRE is limited to how well image captioning models can describe the image. Though captioning models improve over time, we would like to explore novel ways to resolve this bottleneck. For example, Large Multi-Modal Language Models like LLaVa~\cite{liu2023visual} have been shown to perform exceptionally well at generating detailed captions of input images.
    \item The edited images used to personalize text-to-image generation may sometimes be of low quality, leading to poor augmentations in more complex datasets, and we would like to explore ways to resolve this bottleneck. We also acknowledge that the advancement of text-to-image diffusion models to better follow text prompts will eventually lead to performance improvement of ASPIRE.
    \item The different components of ASPIRE add computational overhead to the ASPIRE pipeline (over just the ERM classifier). However, it should be noted that a wealth of literature in offline data augmentation for NLP and CV tasks (through synthetic data generation) almost always employs computationally expensive foundation models for additional data generation. Textual-inversion fine-tuning of diffusion models used in our experiments is also computationally cheaper than full fine-tuning. Lastly, as part of future work, we would like to explore computationally cheaper alternatives to an LLM for information extraction from captions. Additionally, the augmentation process is completely offline and needs to be done just once for each dataset.
    \item We also illustrate some failure cases of ASPIRE in Appendix \ref{sec:failure}.

\end{enumerate}

\section*{Ethics Statement}
\label{sec:ethics}
Image generation models are prone to generating harmful, obscene and offensive context for certain classes pf objects, we prevent this from happening in ASPIRE by using a safety checker for the Stable Diffusion model which estimates whether a generated images could be considered offensive or harmful. For the CelebA dataset, ASPIRE performs modification where genders of people are swapped to debias the model towards certain attributes related a class. This approach is used only to improve the fairness and debiasing of the model.

\bibliography{anthology,custom}

\begin{thebibliography}{68}
\expandafter\ifx\csname natexlab\endcsname\relax\def\natexlab#1{#1}\fi

\bibitem[{Arjovsky et~al.(2019)Arjovsky, Bottou, Gulrajani, and Lopez-Paz}]{arjovsky2019invariant}
Martin Arjovsky, L{\'e}on Bottou, Ishaan Gulrajani, and David Lopez-Paz. 2019.
\newblock Invariant risk minimization.
\newblock \emph{arXiv preprint arXiv:1907.02893}.

\bibitem[{Arpit et~al.(2017)Arpit, Jastrz{\k{e}}bski, Ballas, Krueger, Bengio, Kanwal, Maharaj, Fischer, Courville, Bengio et~al.}]{arpit2017closer}
Devansh Arpit, Stanis{\l}aw Jastrz{\k{e}}bski, Nicolas Ballas, David Krueger, Emmanuel Bengio, Maxinder~S Kanwal, Tegan Maharaj, Asja Fischer, Aaron Courville, Yoshua Bengio, et~al. 2017.
\newblock A closer look at memorization in deep networks.
\newblock In \emph{International conference on machine learning}, pages 233--242. PMLR.

\bibitem[{Asgari et~al.(2022)Asgari, Khani, Khani, Gholami, Tran, Mahdavi~Amiri, and Hamarneh}]{asgari2022masktune}
Saeid Asgari, Aliasghar Khani, Fereshte Khani, Ali Gholami, Linh Tran, Ali Mahdavi~Amiri, and Ghassan Hamarneh. 2022.
\newblock Masktune: Mitigating spurious correlations by forcing to explore.
\newblock \emph{Advances in Neural Information Processing Systems}, 35:23284--23296.

\bibitem[{Azizi et~al.(2023)Azizi, Kornblith, Saharia, Norouzi, and Fleet}]{azizi2023synthetic}
Shekoofeh Azizi, Simon Kornblith, Chitwan Saharia, Mohammad Norouzi, and David~J. Fleet. 2023.
\newblock \href {https://openreview.net/forum?id=DlRsoxjyPm} {Synthetic data from diffusion models improves imagenet classification}.
\newblock \emph{Transactions on Machine Learning Research}.

\bibitem[{Ben-Tal et~al.(2011)Ben-Tal, den Hertog, Waegenaere, Melenberg, and Rennen}]{BenTal2011RobustSO}
Aharon Ben-Tal, Dick den Hertog, Anja~De Waegenaere, Bertrand Melenberg, and Gijs Rennen. 2011.
\newblock \href {https://api.semanticscholar.org/CorpusID:761793} {Robust solutions of optimization problems affected by uncertain probabilities}.
\newblock \emph{Advanced Risk \& Portfolio Management{\textregistered} Research Paper Series}.

\bibitem[{Brendel and Bethge(2019)}]{brendel2019approximating}
Wieland Brendel and Matthias Bethge. 2019.
\newblock Approximating cnns with bag-of-local-features models works surprisingly well on imagenet.
\newblock \emph{arXiv preprint arXiv:1904.00760}.

\bibitem[{Brock et~al.(2018)Brock, Donahue, and Simonyan}]{brock2018large}
Andrew Brock, Jeff Donahue, and Karen Simonyan. 2018.
\newblock Large scale gan training for high fidelity natural image synthesis.
\newblock \emph{arXiv preprint arXiv:1809.11096}.

\bibitem[{Brooks et~al.(2023)Brooks, Holynski, and Efros}]{brooks2023instructpix2pix}
Tim Brooks, Aleksander Holynski, and Alexei~A Efros. 2023.
\newblock Instructpix2pix: Learning to follow image editing instructions.
\newblock In \emph{Proceedings of the IEEE/CVF Conference on Computer Vision and Pattern Recognition}, pages 18392--18402.

\bibitem[{Bruna and Mallat(2013)}]{bruna2013invariant}
Joan Bruna and St{\'e}phane Mallat. 2013.
\newblock Invariant scattering convolution networks.
\newblock \emph{IEEE transactions on pattern analysis and machine intelligence}, 35(8):1872--1886.

\bibitem[{Bruna et~al.(2015)Bruna, Mallat, Bacry, and Muzy}]{bruna2015intermittent}
Joan Bruna, St{\'e}phane Mallat, Emmanuel Bacry, and Jean-Fran{\c{c}}ois Muzy. 2015.
\newblock Intermittent process analysis with scattering moments.

\bibitem[{Deng et~al.(2009)Deng, Dong, Socher, Li, Li, and Fei-Fei}]{deng2009imagenet}
Jia Deng, Wei Dong, Richard Socher, Li-Jia Li, Kai Li, and Li~Fei-Fei. 2009.
\newblock Imagenet: A large-scale hierarchical image database.
\newblock In \emph{2009 IEEE conference on computer vision and pattern recognition}, pages 248--255. Ieee.

\bibitem[{Dunlap et~al.(2023)Dunlap, Mohri, Guillory, Zhang, Darrell, Gonzalez, Raghunathan, and Rohrbach}]{dunlap2023using}
Lisa Dunlap, Clara Mohri, Devin Guillory, Han Zhang, Trevor Darrell, Joseph~E. Gonzalez, Aditi Raghunathan, and Anna Rohrbach. 2023.
\newblock \href {https://openreview.net/forum?id=eR2dG8yjnQ} {Using language to extend to unseen domains}.
\newblock In \emph{The Eleventh International Conference on Learning Representations}.

\bibitem[{Gal et~al.(2023)Gal, Alaluf, Atzmon, Patashnik, Bermano, Chechik, and Cohen-or}]{gal2023an}
Rinon Gal, Yuval Alaluf, Yuval Atzmon, Or~Patashnik, Amit~Haim Bermano, Gal Chechik, and Daniel Cohen-or. 2023.
\newblock \href {https://openreview.net/forum?id=NAQvF08TcyG} {An image is worth one word: Personalizing text-to-image generation using textual inversion}.
\newblock In \emph{The Eleventh International Conference on Learning Representations}.

\bibitem[{Geirhos et~al.(2020)Geirhos, Jacobsen, Michaelis, Zemel, Brendel, Bethge, and Wichmann}]{Geirhos_2020}
Robert Geirhos, Jörn-Henrik Jacobsen, Claudio Michaelis, Richard Zemel, Wieland Brendel, Matthias Bethge, and Felix~A. Wichmann. 2020.
\newblock \href {https://doi.org/10.1038/s42256-020-00257-z} {Shortcut learning in deep neural networks}.
\newblock \emph{Nature Machine Intelligence}, 2(11):665--673.

\bibitem[{Gowal et~al.(2021)Gowal, Rebuffi, Wiles, Stimberg, Calian, and Mann}]{gowal2021improving}
Sven Gowal, Sylvestre-Alvise Rebuffi, Olivia Wiles, Florian Stimberg, Dan~Andrei Calian, and Timothy~A Mann. 2021.
\newblock Improving robustness using generated data.
\newblock \emph{Advances in Neural Information Processing Systems}, 34:4218--4233.

\bibitem[{He et~al.(2022)He, Sun, Yu, Xue, Zhang, Torr, Bai, and Qi}]{he2022synthetic}
Ruifei He, Shuyang Sun, Xin Yu, Chuhui Xue, Wenqing Zhang, Philip Torr, Song Bai, and Xiaojuan Qi. 2022.
\newblock Is synthetic data from generative models ready for image recognition?
\newblock \emph{arXiv preprint arXiv:2210.07574}.

\bibitem[{Hermann and Lampinen(2020)}]{10.5555/3495724.3496562}
Katherine~L. Hermann and Andrew~K. Lampinen. 2020.
\newblock What shapes feature representations? exploring datasets, architectures, and training.
\newblock In \emph{Proceedings of the 34th International Conference on Neural Information Processing Systems}, NIPS'20, Red Hook, NY, USA. Curran Associates Inc.

\bibitem[{Hu et~al.(2018)Hu, Niu, Sato, and Sugiyama}]{HuNSS18}
Weihua Hu, Gang Niu, Issei Sato, and Masashi Sugiyama. 2018.
\newblock \href {http://proceedings.mlr.press/v80/hu18a.html} {Does distributionally robust supervised learning give robust classifiers?}
\newblock In \emph{Proceedings of the 35th International Conference on Machine Learning, ICML 2018, Stockholmsmässan, Stockholm, Sweden, July 10-15, 2018}, volume~80 of \emph{JMLR Workshop and Conference Proceedings}, pages 2034--2042. JMLR.org.

\bibitem[{Idrissi et~al.(2022)Idrissi, Arjovsky, Pezeshki, and Lopez-Paz}]{idrissi2022simple}
Badr~Youbi Idrissi, Martin Arjovsky, Mohammad Pezeshki, and David Lopez-Paz. 2022.
\newblock Simple data balancing achieves competitive worst-group-accuracy.
\newblock In \emph{Conference on Causal Learning and Reasoning}, pages 336--351. PMLR.

\bibitem[{Jabri et~al.(2016)Jabri, Joulin, and Van Der~Maaten}]{jabri2016revisiting}
Allan Jabri, Armand Joulin, and Laurens Van Der~Maaten. 2016.
\newblock Revisiting visual question answering baselines.
\newblock In \emph{European conference on computer vision}, pages 727--739. Springer.

\bibitem[{Jacobsen et~al.(2019)Jacobsen, Behrmann, Zemel, and Bethge}]{jacobsen2018excessive}
Joern-Henrik Jacobsen, Jens Behrmann, Richard Zemel, and Matthias Bethge. 2019.
\newblock \href {https://openreview.net/forum?id=BkfbpsAcF7} {Excessive invariance causes adversarial vulnerability}.
\newblock In \emph{International Conference on Learning Representations}.

\bibitem[{Joshi et~al.(2023)Joshi, Yang, Xue, Yang, and Mirzasoleiman}]{joshi2023towards}
Siddharth Joshi, Yu~Yang, Yihao Xue, Wenhan Yang, and Baharan Mirzasoleiman. 2023.
\newblock Towards mitigating spurious correlations in the wild: A benchmark \& a more realistic dataset.
\newblock \emph{arXiv preprint arXiv:2306.11957}.

\bibitem[{Kalimeris et~al.(2019)Kalimeris, Kaplun, Nakkiran, Edelman, Yang, Barak, and Zhang}]{kalimeris2019sgd}
Dimitris Kalimeris, Gal Kaplun, Preetum Nakkiran, Benjamin Edelman, Tristan Yang, Boaz Barak, and Haofeng Zhang. 2019.
\newblock Sgd on neural networks learns functions of increasing complexity.
\newblock \emph{Advances in neural information processing systems}, 32.

\bibitem[{Khani and Liang(2021)}]{khani2021removing}
Fereshte Khani and Percy Liang. 2021.
\newblock Removing spurious features can hurt accuracy and affect groups disproportionately.
\newblock In \emph{Proceedings of the 2021 ACM conference on fairness, accountability, and transparency}, pages 196--205.

\bibitem[{Kim et~al.(2023)Kim, Koepke, Schmid, and Akata}]{kim2023exposing}
Jae~Myung Kim, A~Koepke, Cordelia Schmid, and Zeynep Akata. 2023.
\newblock Exposing and mitigating spurious correlations for cross-modal retrieval.
\newblock In \emph{Proceedings of the IEEE/CVF Conference on Computer Vision and Pattern Recognition}, pages 2584--2594.

\bibitem[{Kirichenko et~al.(2023)Kirichenko, Izmailov, and Wilson}]{kirichenko2023last}
Polina Kirichenko, Pavel Izmailov, and Andrew~Gordon Wilson. 2023.
\newblock \href {https://openreview.net/forum?id=Zb6c8A-Fghk} {Last layer re-training is sufficient for robustness to spurious correlations}.
\newblock In \emph{The Eleventh International Conference on Learning Representations}.

\bibitem[{Kirillov et~al.(2023)Kirillov, Mintun, Ravi, Mao, Rolland, Gustafson, Xiao, Whitehead, Berg, Lo et~al.}]{kirillov2023segment}
Alexander Kirillov, Eric Mintun, Nikhila Ravi, Hanzi Mao, Chloe Rolland, Laura Gustafson, Tete Xiao, Spencer Whitehead, Alexander~C Berg, Wan-Yen Lo, et~al. 2023.
\newblock Segment anything.
\newblock \emph{arXiv preprint arXiv:2304.02643}.

\bibitem[{Kong et~al.(2023)Kong, Yuan, Hao, and Henao}]{kong2023mitigating}
Fanjie Kong, Shuai Yuan, Weituo Hao, and Ricardo Henao. 2023.
\newblock Mitigating test-time bias for fair image retrieval.
\newblock \emph{arXiv preprint arXiv:2305.19329}.

\bibitem[{Lee et~al.(2022)Lee, Yao, and Finn}]{DBLP:journals/corr/abs-2202-03418}
Yoonho Lee, Huaxiu Yao, and Chelsea Finn. 2022.
\newblock \href {http://arxiv.org/abs/2202.03418} {Diversify and disambiguate: Learning from underspecified data}.
\newblock \emph{CoRR}, abs/2202.03418.

\bibitem[{Li et~al.(2022)Li, Ling, Kim, Kreis, Fidler, and Torralba}]{li2022bigdatasetgan}
Daiqing Li, Huan Ling, Seung~Wook Kim, Karsten Kreis, Sanja Fidler, and Antonio Torralba. 2022.
\newblock Bigdatasetgan: Synthesizing imagenet with pixel-wise annotations.
\newblock In \emph{Proceedings of the IEEE/CVF Conference on Computer Vision and Pattern Recognition}, pages 21330--21340.

\bibitem[{Liu et~al.(2021{\natexlab{a}})Liu, Haghgoo, Chen, Raghunathan, Koh, Sagawa, Liang, and Finn}]{pmlr-v139-liu21f}
Evan~Z Liu, Behzad Haghgoo, Annie~S Chen, Aditi Raghunathan, Pang~Wei Koh, Shiori Sagawa, Percy Liang, and Chelsea Finn. 2021{\natexlab{a}}.
\newblock \href {https://proceedings.mlr.press/v139/liu21f.html} {Just train twice: Improving group robustness without training group information}.
\newblock In \emph{Proceedings of the 38th International Conference on Machine Learning}, volume 139 of \emph{Proceedings of Machine Learning Research}, pages 6781--6792. PMLR.

\bibitem[{Liu et~al.(2021{\natexlab{b}})Liu, Haghgoo, Chen, Raghunathan, Koh, Sagawa, Liang, and Finn}]{liu2021just}
Evan~Zheran Liu, Behzad Haghgoo, Annie~S. Chen, Aditi Raghunathan, Pang~Wei Koh, Shiori Sagawa, Percy Liang, and Chelsea Finn. 2021{\natexlab{b}}.
\newblock \href {http://arxiv.org/abs/2107.09044} {Just train twice: Improving group robustness without training group information}.

\bibitem[{Liu et~al.(2023{\natexlab{a}})Liu, Li, Wu, and Lee}]{liu2023llava}
Haotian Liu, Chunyuan Li, Qingyang Wu, and Yong~Jae Lee. 2023{\natexlab{a}}.
\newblock Visual instruction tuning.

\bibitem[{Liu et~al.(2023{\natexlab{b}})Liu, Li, Wu, and Lee}]{liu2023visual}
Haotian Liu, Chunyuan Li, Qingyang Wu, and Yong~Jae Lee. 2023{\natexlab{b}}.
\newblock Visual instruction tuning.
\newblock \emph{arXiv preprint arXiv:2304.08485}.

\bibitem[{Liu et~al.(2023{\natexlab{c}})Liu, Zeng, Ren, Li, Zhang, Yang, Li, Yang, Su, Zhu et~al.}]{liu2023grounding}
Shilong Liu, Zhaoyang Zeng, Tianhe Ren, Feng Li, Hao Zhang, Jie Yang, Chunyuan Li, Jianwei Yang, Hang Su, Jun Zhu, et~al. 2023{\natexlab{c}}.
\newblock Grounding dino: Marrying dino with grounded pre-training for open-set object detection.
\newblock \emph{arXiv preprint arXiv:2303.05499}.

\bibitem[{Liu et~al.(2023{\natexlab{d}})Liu, Li, and Lin}]{liu2023cross}
Yang Liu, Guanbin Li, and Liang Lin. 2023{\natexlab{d}}.
\newblock Cross-modal causal relational reasoning for event-level visual question answering.
\newblock \emph{IEEE Transactions on Pattern Analysis and Machine Intelligence}.

\bibitem[{Liu et~al.(2015)Liu, Luo, Wang, and Tang}]{liu2015deep}
Ziwei Liu, Ping Luo, Xiaogang Wang, and Xiaoou Tang. 2015.
\newblock Deep learning face attributes in the wild.
\newblock In \emph{Proceedings of the IEEE international conference on computer vision}, pages 3730--3738.

\bibitem[{Moayeri et~al.(2022)Moayeri, Singla, and Feizi}]{moayeri2022hard}
Mazda Moayeri, Sahil Singla, and Soheil Feizi. 2022.
\newblock Hard imagenet: Segmentations for objects with strong spurious cues.
\newblock \emph{Advances in Neural Information Processing Systems}, 35:10068--10077.

\bibitem[{Nagarajan et~al.(2021)Nagarajan, Andreassen, and Neyshabur}]{nagarajan2021understanding}
Vaishnavh Nagarajan, Anders Andreassen, and Behnam Neyshabur. 2021.
\newblock \href {https://openreview.net/forum?id=fSTD6NFIW_b} {Understanding the failure modes of out-of-distribution generalization}.
\newblock In \emph{International Conference on Learning Representations}.

\bibitem[{Nam et~al.(2020)Nam, Cha, Ahn, Lee, and Shin}]{nam2020learning}
Junhyun Nam, Hyuntak Cha, Sungsoo Ahn, Jaeho Lee, and Jinwoo Shin. 2020.
\newblock Learning from failure: De-biasing classifier from biased classifier.
\newblock \emph{Advances in Neural Information Processing Systems}, 33:20673--20684.

\bibitem[{Nichol et~al.(2021)Nichol, Dhariwal, Ramesh, Shyam, Mishkin, McGrew, Sutskever, and Chen}]{nichol2021glide}
Alex Nichol, Prafulla Dhariwal, Aditya Ramesh, Pranav Shyam, Pamela Mishkin, Bob McGrew, Ilya Sutskever, and Mark Chen. 2021.
\newblock Glide: Towards photorealistic image generation and editing with text-guided diffusion models.
\newblock \emph{arXiv preprint arXiv:2112.10741}.

\bibitem[{OpenAI(2023)}]{openai2023gpt4}
OpenAI. 2023.
\newblock \href {http://arxiv.org/abs/2303.08774} {Gpt-4 technical report}.

\bibitem[{Oren et~al.(2019)Oren, Sagawa, Hashimoto, and Liang}]{oren-etal-2019-distributionally}
Yonatan Oren, Shiori Sagawa, Tatsunori~B. Hashimoto, and Percy Liang. 2019.
\newblock \href {https://doi.org/10.18653/v1/D19-1432} {Distributionally robust language modeling}.
\newblock In \emph{Proceedings of the 2019 Conference on Empirical Methods in Natural Language Processing and the 9th International Joint Conference on Natural Language Processing (EMNLP-IJCNLP)}, pages 4227--4237, Hong Kong, China. Association for Computational Linguistics.

\bibitem[{Pezeshki et~al.(2021)Pezeshki, Kaba, Bengio, Courville, Precup, and Lajoie}]{pezeshki2021gradient}
Mohammad Pezeshki, S{\'e}kou-Oumar Kaba, Yoshua Bengio, Aaron Courville, Doina Precup, and Guillaume Lajoie. 2021.
\newblock \href {https://openreview.net/forum?id=aExAsh1UHZo} {Gradient starvation: A learning proclivity in neural networks}.
\newblock In \emph{Advances in Neural Information Processing Systems}.

\bibitem[{Prabhu et~al.(2023)Prabhu, Yenamandra, Chattopadhyay, and Hoffman}]{prabhu2023lance}
Viraj~Uday Prabhu, Sriram Yenamandra, Prithvijit Chattopadhyay, and Judy Hoffman. 2023.
\newblock \href {https://openreview.net/forum?id=BbIxB4xnbq} {{LANCE}: Stress-testing visual models by generating language-guided counterfactual images}.
\newblock In \emph{Thirty-seventh Conference on Neural Information Processing Systems}.

\bibitem[{Rahaman et~al.(2019)Rahaman, Baratin, Arpit, Draxler, Lin, Hamprecht, Bengio, and Courville}]{rahaman2019on}
Nasim Rahaman, Aristide Baratin, Devansh Arpit, Felix Draxler, Min Lin, Fred Hamprecht, Yoshua Bengio, and Aaron Courville. 2019.
\newblock \href {https://openreview.net/forum?id=r1gR2sC9FX} {On the spectral bias of neural networks}.

\bibitem[{Sagawa et~al.(2019)Sagawa, Koh, Hashimoto, and Liang}]{sagawa2019distributionally}
Shiori Sagawa, Pang~Wei Koh, Tatsunori~B Hashimoto, and Percy Liang. 2019.
\newblock Distributionally robust neural networks for group shifts: On the importance of regularization for worst-case generalization.
\newblock \emph{arXiv preprint arXiv:1911.08731}.

\bibitem[{Sagawa* et~al.(2020)Sagawa*, Koh*, Hashimoto, and Liang}]{Sagawa*2020Distributionally}
Shiori Sagawa*, Pang~Wei Koh*, Tatsunori~B. Hashimoto, and Percy Liang. 2020.
\newblock \href {https://openreview.net/forum?id=ryxGuJrFvS} {Distributionally robust neural networks}.
\newblock In \emph{International Conference on Learning Representations}.

\bibitem[{Sagawa et~al.(2020)Sagawa, Koh, Hashimoto, and Liang}]{sagawa2020distributionally}
Shiori Sagawa, Pang~Wei Koh, Tatsunori~B. Hashimoto, and Percy Liang. 2020.
\newblock \href {http://arxiv.org/abs/1911.08731} {Distributionally robust neural networks for group shifts: On the importance of regularization for worst-case generalization}.

\bibitem[{Shah et~al.(2020)Shah, Tamuly, Raghunathan, Jain, and Netrapalli}]{shah2020pitfalls}
Harshay Shah, Kaustav Tamuly, Aditi Raghunathan, Prateek Jain, and Praneeth Netrapalli. 2020.
\newblock The pitfalls of simplicity bias in neural networks.
\newblock \emph{Advances in Neural Information Processing Systems}, 33.

\bibitem[{Shinoda et~al.(2023)Shinoda, Sugawara, and Aizawa}]{shinoda2023shortcut}
Kazutoshi Shinoda, Saku Sugawara, and Akiko Aizawa. 2023.
\newblock Which shortcut solution do question answering models prefer to learn?
\newblock In \emph{Proceedings of the AAAI Conference on Artificial Intelligence}, volume~37, pages 13564--13572.

\bibitem[{Suvorov et~al.(2022)Suvorov, Logacheva, Mashikhin, Remizova, Ashukha, Silvestrov, Kong, Goka, Park, and Lempitsky}]{suvorov2022resolution}
Roman Suvorov, Elizaveta Logacheva, Anton Mashikhin, Anastasia Remizova, Arsenii Ashukha, Aleksei Silvestrov, Naejin Kong, Harshith Goka, Kiwoong Park, and Victor Lempitsky. 2022.
\newblock Resolution-robust large mask inpainting with fourier convolutions.
\newblock In \emph{Proceedings of the IEEE/CVF winter conference on applications of computer vision}, pages 2149--2159.

\bibitem[{Taghanaki et~al.(2022)Taghanaki, Khani, Khani, Gholami, Tran, Mahdavi-Amiri, and Hamarneh}]{taghanaki2022masktune}
Saeid~Asgari Taghanaki, Aliasghar Khani, Fereshte Khani, Ali Gholami, Linh Tran, Ali Mahdavi-Amiri, and Ghassan Hamarneh. 2022.
\newblock \href {http://arxiv.org/abs/2210.00055} {Masktune: Mitigating spurious correlations by forcing to explore}.

\bibitem[{Tong et~al.(2024)Tong, Jones, and Steinhardt}]{tong2024mass}
Shengbang Tong, Erik Jones, and Jacob Steinhardt. 2024.
\newblock Mass-producing failures of multimodal systems with language models.
\newblock \emph{Advances in Neural Information Processing Systems}, 36.

\bibitem[{Torralba and Efros(2011)}]{torralba2011unbiased}
Antonio Torralba and Alexei~A Efros. 2011.
\newblock Unbiased look at dataset bias.
\newblock In \emph{CVPR 2011}, pages 1521--1528. IEEE.

\bibitem[{Touvron et~al.(2023)Touvron, Martin, Stone, Albert, Almahairi, Babaei, Bashlykov, Batra, Bhargava, Bhosale et~al.}]{touvron2023llama}
Hugo Touvron, Louis Martin, Kevin Stone, Peter Albert, Amjad Almahairi, Yasmine Babaei, Nikolay Bashlykov, Soumya Batra, Prajjwal Bhargava, Shruti Bhosale, et~al. 2023.
\newblock Llama 2: Open foundation and fine-tuned chat models.
\newblock \emph{arXiv preprint arXiv:2307.09288}.

\bibitem[{Trabucco et~al.(2023)Trabucco, Doherty, Gurinas, and Salakhutdinov}]{trabucco2023effective}
Brandon Trabucco, Kyle Doherty, Max Gurinas, and Ruslan Salakhutdinov. 2023.
\newblock Effective data augmentation with diffusion models.
\newblock \emph{arXiv preprint arXiv:2302.07944}.

\bibitem[{Valle-Perez et~al.(2018)Valle-Perez, Camargo, and Louis}]{valle2018deep}
Guillermo Valle-Perez, Chico~Q Camargo, and Ard~A Louis. 2018.
\newblock Deep learning generalizes because the parameter-function map is biased towards simple functions.
\newblock \emph{arXiv preprint arXiv:1805.08522}.

\bibitem[{Vendrow et~al.(2023)Vendrow, Jain, Engstrom, and Madry}]{vendrow2023dataset}
Joshua Vendrow, Saachi Jain, Logan Engstrom, and Aleksander Madry. 2023.
\newblock \href {http://arxiv.org/abs/2302.07865} {Dataset interfaces: Diagnosing model failures using controllable counterfactual generation}.

\bibitem[{Wah et~al.(2011)Wah, Branson, Welinder, Perona, and Belongie}]{wah2011caltech}
Catherine Wah, Steve Branson, Peter Welinder, Pietro Perona, and Serge Belongie. 2011.
\newblock The caltech-ucsd birds-200-2011 dataset.

\bibitem[{Wang et~al.(2022)Wang, Yang, Hu, Li, Lin, Gan, Liu, Liu, and Wang}]{wang2022git}
Jianfeng Wang, Zhengyuan Yang, Xiaowei Hu, Linjie Li, Kevin Lin, Zhe Gan, Zicheng Liu, Ce~Liu, and Lijuan Wang. 2022.
\newblock Git: A generative image-to-text transformer for vision and language.
\newblock \emph{arXiv preprint arXiv:2205.14100}.

\bibitem[{Welinder et~al.(2010)Welinder, Branson, Mita, Wah, Schroff, Belongie, and Perona}]{399}
Peter Welinder, Steve Branson, Takeshi Mita, Catherine Wah, Florian Schroff, Serge Belongie, and Pietro Perona. 2010.
\newblock \href {/se3/wp-content/uploads/2014/09/WelinderEtal10_CUB-200.pdf, http://www.vision.caltech.edu/visipedia/CUB-200.html} {Caltech-ucsd birds 200}.
\newblock Technical Report CNS-TR-201, Caltech.

\bibitem[{Wiles et~al.(2022)Wiles, Albuquerque, and Gowal}]{wiles2022discovering}
Olivia Wiles, Isabela Albuquerque, and Sven Gowal. 2022.
\newblock \href {https://openreview.net/forum?id=maBZZ_W0lD} {Discovering bugs in vision models using off-the-shelf image generation and captioning}.
\newblock In \emph{NeurIPS ML Safety Workshop}.

\bibitem[{Wilson et~al.(2017)Wilson, Roelofs, Stern, Srebro, and Recht}]{wilson2017marginal}
Ashia~C Wilson, Rebecca Roelofs, Mitchell Stern, Nati Srebro, and Benjamin Recht. 2017.
\newblock The marginal value of adaptive gradient methods in machine learning.
\newblock \emph{Advances in neural information processing systems}, 30.

\bibitem[{Yang et~al.(2023)Yang, Nushi, Palangi, and Mirzasoleiman}]{yang2023mitigating}
Yu~Yang, Besmira Nushi, Hamid Palangi, and Baharan Mirzasoleiman. 2023.
\newblock Mitigating spurious correlations in multi-modal models during fine-tuning.
\newblock \emph{arXiv preprint arXiv:2304.03916}.

\bibitem[{Zhang et~al.(2021)Zhang, Menon, Veit, Bhojanapalli, Kumar, and Sra}]{zhang2021coping}
Jingzhao Zhang, Aditya~Krishna Menon, Andreas Veit, Srinadh Bhojanapalli, Sanjiv Kumar, and Suvrit Sra. 2021.
\newblock \href {https://openreview.net/forum?id=BtZhsSGNRNi} {Coping with label shift via distributionally robust optimisation}.
\newblock In \emph{International Conference on Learning Representations}.

\bibitem[{Zhang et~al.(2022)Zhang, Sohoni, Zhang, Finn, and R{\'e}}]{zhang2022correct}
Michael Zhang, Nimit~S Sohoni, Hongyang~R Zhang, Chelsea Finn, and Christopher R{\'e}. 2022.
\newblock Correct-n-contrast: A contrastive approach for improving robustness to spurious correlations.
\newblock \emph{arXiv preprint arXiv:2203.01517}.

\bibitem[{Zhou et~al.(2017)Zhou, Lapedriza, Khosla, Oliva, and Torralba}]{zhou2017places}
Bolei Zhou, Agata Lapedriza, Aditya Khosla, Aude Oliva, and Antonio Torralba. 2017.
\newblock Places: A 10 million image database for scene recognition.
\newblock \emph{IEEE transactions on pattern analysis and machine intelligence}, 40(6):1452--1464.

\end{thebibliography}

\appendix

\section{Appendix}
\label{sec:appendix}
\subsection{Dataset Details}
\label{subsec:dataset_details}
\label{tab:dataset_details}
Table \ref{tab:dataset_details} shows dataset details for all 4 datasets used in our experiments. As clearly visible, there is a notable disparity between the number of images representing minority groups (non-spurious images) and those representing majority groups (images with spuriously correlated features). In contrast, the test set for each dataset maintains a balanced representation between the minority and majority groups. This can lead classifiers to quickly adopt spurious correlations, resulting in sub-optimal performance on the test set.
% \vspace{-1em}

\begin{table}[h!]
    \center
\resizebox{0.99\columnwidth}{!}{
\begin{tabular}{lccccc}
\toprule
\multicolumn{1}{c}{Dataset} &
  \multicolumn{2}{c}{Train} &
  \multicolumn{2}{c}{Test} \\ 
  \cmidrule(lr){2-3} \cmidrule(lr){4-5} &
  Majority &
  Minority &
  Majority &
  Minority \\
  & Group & Group & Group & Group \\ \midrule
  
Hard ImageNet & 19097 & 0 & 375 & 375\\

Waterbirds & 4555 & 240 &  2897 & 2897\\

CelebA & 161383 & 1387 & 19782 & 180\\ 

Spuco Dogs & 17000 & 1000 & 1000 & 1000\\ 
\bottomrule
\end{tabular}
}

\caption{Dataset details}
\end{table}

% \vspace{-1.25em}
\subsection{Algorithm}
\label{sec:algorithm}
Algorithm \ref{algo:algorithm} describes algorithmically the ASPIRE pipeline. Readers can refer to the algorithm for a detailed step-by-step understanding of the workings of ASPIRE.
\vspace{0.5mm}

\subsection{Traditional NLP algorithm details}

\noindent \textbf{Introduction}
Extracting foreground objects and the background from a caption using traditional Natural Language Processing (NLP) techniques and libraries like SpaCy involves several steps. Here's a general approach:

\noindent \textbf{Text Preprocessing}
First, preprocess the text to ensure it's in a suitable format for analysis. This might include:
\begin{itemize}
    \item Lowercasing all words.
    \item Removing punctuation and special characters.
    \item Tokenization: Breaking the text into individual words (tokens).
\end{itemize}

\noindent \textbf{Part-of-Speech Tagging}
Use SpaCy to perform part-of-speech (POS) tagging, which identifies the grammatical parts of speech for each word (e.g., noun, verb, adjective). This is crucial for identifying potential objects and elements of the scene.

\noindent \textbf{Named Entity Recognition (NER)}
Employ Named Entity Recognition to identify named entities in the text, which can include names of people, places, organizations, or other proper nouns. These entities can be part of the foreground or background.

\noindent \textbf{Dependency Parsing}
Dependency parsing helps understand the grammatical structure of the sentence, showing how words relate to each other. This is useful to distinguish between main subjects (likely foreground objects) and contextual elements (possibly background).

\noindent \textbf{Chunking or Phrase Detection}
Use chunking or phrase detection to group together contiguous sequences of tokens that form meaningful phrases. Noun phrases, in particular, are often key in identifying objects and scene elements.

\noindent \textbf{Identifying Foreground and Background}
\noindent \textbf{Foreground Objects}
Typically, these are nouns or noun phrases that are the main subjects or objects of the sentence. They often appear with adjectives and are part of active clauses.

\noindent \textbf{Background Information}
This can include descriptions of settings, locations, or contexts. Adverbial phrases and clauses, as well as descriptive language, can signal background details.

\noindent \textbf{SpaCy Implementation}
Here's a simple implementation using SpaCy:

\begin{spverbatim}
import spacy

# Load the SpaCy model
nlp = spacy.load("en_core_web_sm")

def extract_foreground_background(text):
    doc = nlp(text)

    foreground = []
    background = []

    for token in doc:
        # Check for nouns and proper nouns for foreground
        if token.pos_ in ["NOUN", "PROPN"]:
            foreground.append(token.text)

        # Background might be set by adverbial phrases or adjectives
        if token.pos_ in ["ADJ", "ADV"]:
            background.append(token.text)

        # Check for named entities
        if token.ent_type_:
            if token.ent_type_ in ["PERSON", "ORG", "GPE"]:
                foreground.append(token.text)
            else:
                background.append(token.text)

    return foreground, background

# Example Usage
text = "The cat sat on the mat in the sunny room."
foreground, background = extract_foreground_background(text)
print("Foreground:", foreground)
print("Background:", background)
\end{spverbatim}

\renewcommand{\algorithmiccomment}[1]{\hfill$\triangleright${#1}}
\newcommand{\myComment}[1]{\textcolor{blue}{\ttfamily{// #1}}}
\begin{algorithm*}[h!]
\KwData{Image Classification Dataset $\mathcal{D}_{train} \rightarrow \{x_{i} \ (Image), y_{i} \ (Label)\}$;}
$\mathcal{E}$ = $Classifier(x_{i}, y_{i})$ \myComment{Image classification model}\\
$\mathcal{C} = \operatorname{Captioning}(x_{i})$ \myComment{Image captioning model}\\
$\mathcal{L} = \operatorname{LLM}(Prompt, Captions, y)$  \myComment{LLM to extract foreground and background objects.}\\
$\mathcal{BG} = \operatorname{InstructPix2Pix}(b,\tilde{b},x_{i})$ \myComment{InstructPix2Pix to convert background of an image.}\\
$\mathcal{G} = \operatorname{GroundingDino}(f,x_{i})$ \myComment{Creates bounding box around objects.}\\
$\mathcal{S} = \operatorname{SegmentAnything}(bb)$ \myComment{Extracts image segmentation maps from bounding boxes.}\\
$\mathcal{I} = \operatorname{InpaintAnything}(\mathcal{M},x_{i})$ \myComment{Removal of objects corresponding to segmentation maps.}\\
$\mathcal{D}_{correct} \leftarrow \emptyset$
\For{$x_i$ in $\mathcal{D}_{train}$}{
 \myComment{Consider only the images which are predicted correctly.}\\
    \If{$\mathcal{E}(x_{i}) == y_{i}$}{
       $\mathcal{D}_{correct} \leftarrow \mathcal{D}_{correct} \cup \{(x_{i}, y_{i})\}$;
    }
}
Sample $p\%$ of $\mathcal{D}_{correct}$ to create $\mathcal{D}_{hold}$.\\
$\mathcal{D}_{captions} \leftarrow \operatorname{\mathcal{C}}(\mathcal{D}_{hold})$ \myComment{Caption the images in the holdout set.}\\
$\mathcal{F}_{i},\mathcal{B}_{i} \leftarrow \operatorname{\mathcal{L}}(Prompt, \mathcal{D}_{captions}, \mathcal{D}^{y}_{hold})$ \myComment{Extract foreground and background objects by prompting the LLM.}\\
$\mathcal{D}_{synth}\leftarrow\emptyset, \mathcal{T}_{synth}\leftarrow\emptyset$;\\
\For{$\{f\}$ in $\mathcal{F}_{i}$}{
    $bb \leftarrow \operatorname{\mathcal{G}}(f,x_{i})$; \myComment{Create the bounding boxes.} \\
    $\mathcal{M} \leftarrow \operatorname{\mathcal{S}}(bb)$; \myComment{Extract the segmentation maps.}\\
    $x^{mod}_{i} \leftarrow \operatorname{\mathcal{I}}(\mathcal{M})$; \myComment{Modify image by removing the foreground object.}\\
    \myComment{Consider only the images which are predicted wrong after modification.}\\
    \If{$y^{correct}_{i} \neq \operatorname{\mathcal{E}}(x^{mod}_{i})$}{
        $\mathcal{D}_{synth} \leftarrow \mathcal{D}_{synth} \cup \{x^{mod}_{i}\}$;\\
        $\mathcal{T}_{synth} \leftarrow \mathcal{T}_{synth} \cup \{f\}$;
    }
}
\For{$\{b, \tilde{b}\}$ in $\mathcal{B}_{i}$}{
    $x^{mod}_{i} \leftarrow \operatorname{\mathcal{BG}}(b,\tilde{b},x_{i})$;\myComment{Change image background as suggested by the LLM.}\\
     \myComment{Consider only the images which are predicted wrong after modification.}\\
    \If{$y^{correct}_{i} \neq \operatorname{\mathcal{E}}(x^{mod}_{i})$}{
        $\mathcal{D}_{synth} \leftarrow \mathcal{D}_{synth} \cup \{x^{mod}_{i}\}$;\\
        $\mathcal{T}_{synth} \leftarrow \mathcal{T}_{synth} \cup \{b\}$;
    }
}
\myComment{Collapse synthetic dataset based on text phrases that are similar to each other. Select top-k items that have the highest count per image class in the dataset.}\\
$\mathcal{T}^{k}_{synth},\mathcal{D}^{k}_{synth} \leftarrow \operatorname{TopK}(\operatorname{Col}(\mathcal{T}_{synth}, \mathcal{D}_{synth}))$\\
Train the Stable Diffusion model $\mathcal{SD}$ using $\mathcal{D}^{k}_{synth}$.\\
Generate $\mathcal{D}_{aug}$ from $\mathcal{SD}$. \\
\myComment{Creating a new training dataset by combining the augmentations with the original training data.}\\
$\mathcal{D}^{new}_{train} \leftarrow \mathcal{D}_{train} \cup \mathcal{D}_{aug}$; \\
\myComment{Retrain the original image captioning model on the new training data.}\\
Retrain $\mathcal{E}$ on $\mathcal{D}^{new}_{train}$.\\
\caption{ASPIRE Data Augmentation Algorithm}
\label{algo:algorithm}
\end{algorithm*}

\subsection{Details on Baselines}
\label{subsec:details}

To maintain training efficiency, for training each baseline with ASPIRE augmentations, we add only 1$\times$ more augmentations to the original dataset for CelebA, Waterbirds, and SPUCO Dogs, or effectively or effectively double the number of non-spurious minority group images in each dataset. These 3 datasets have labeled minority groups, and thus, the number of augmentations to be added amounted to the total minority group images in each class of the original dataset. For Hard ImageNet, we add as many more augmentations as the total number of original training samples in each class of the original dataset. We elaborate on the rationale behind the choice of our baseline setup in Appendix \ref{subsec:appendix_baselines}, where we also describe why we choose not to compare ASPIRE with large multi-modal models. We next describe how we add ASPIRE augmentations to the original training pipeline for different baselines.
\vspace{0.5mm}

{\noindent \textbf{Emperical Risk Minimization}} (ERM) For this baseline, we compare a ResNet-50 model trained using ERM (with SGD) on the original dataset with a ResNet-50 model trained on the original dataset augmented with ASPIRE augmentations. For ERM, we just add ASPIRE augmentations to the initial training set.
\vspace{0.5mm}

{\noindent \textbf{1-stage training baselines.}} \textbf{Group DRO} \cite{sagawa2019distributionally} is a state-of-the-art method that uses group information on train and adaptively upweights worst-group examples during training. \textbf{SUBG} \cite{idrissi2022simple} is ERM applied to a random subset of the data where the groups are equally represented, which was recently shown to be a strong baseline. We also add ASPIRE augmentations to the initial training set for both baselines.
% \vspace{0.5mm}

{\noindent \textbf{2-stage training baselines.}} \textbf{Just Train Twice} (JTT). JTT follows a 2-stage training process wherein they first identify training examples that are misclassified by a standard ERM model and then train the final model by upweighting the examples identified in the first stage. \textbf{Learning from Failure} (LfF). \cite{nam2020learning} Similar to JTT, LfF follows a 2-stage training process wherein they first identify training examples that are misclassified by a biased ERM model and then train the final model by re-weight training samples using the relative difficulty score based on the loss of the biased model. \textbf{Correct-n-Contrast} (CnC) \cite{zhang2022correct} detects the minority group examples similarly to JTT and uses a contrastive objective to learn representations robust to spurious correlations. \textbf{Deep Feature Reweighting} (DFR). \cite{kirichenko2023last} DFR follows a 2-stage training process wherein they first fine-tune a pre-trained ResNet model (pre-trained on the entire ImageNet dataset) using ERM on the entire train split followed by re-training the last layer using a small set from the train with an equal number of instances for both majority and minority groups. \textbf{MaskTune.} \cite{asgari2022masktune} follows a 2-stage training process, wherein they first fine-tune a ResNet model on a dataset using ERM on the entire train split followed by re-training the model with new masked data for one full epoch. For all these baselines, we add ASPIRE augmentations to the set used in the second stage of training.

\subsection{Choice of Baselines}
\label{subsec:appendix_baselines}
To the best of our knowledge, there exists no prior method in literature that generates minority group images to expand the training set. Most work has focused on devising novel training methods for robust classification, all of which are complementary to ASPIRE and compared to our method in this paper. As also mentioned in Section~\ref{sec:literature} of our paper, generative data augmentation for improving overall accuracy has been explored but is unrelated to our method. Additionally, the primary aim of ASPIRE is to improve the downstream performance of \textit{image classification models}. We acknowledge that other types of models, like instruction-tuned Vision-Language Models~\cite{liu2023visual}, might identify and classify the image correctly into a predefined class given specific prompts (again, this is an underexplored area in CV), but comparing this is beyond the scope of this paper and experimental setting. Our setting is consistent with most prior art in  (methods listed in Table~\ref{tab:main}).
% \vspace{-1em}

\subsection{Prompts}
\label{subsec:llama}

{\noindent \textbf{GPT-4.}}The general-purpose prompt we use for GPT-4 is listed as follows: \textit{I will provide you with a list of tuples. Each tuple in the list has 2 items: the first is a caption of an image and the second is the label of the image. For each, you will have to return a JSON with 3 lists. One list should be the list of all phrases from the caption that are objects that appear in the foreground of the image but ignore objects that correspond to the actual label (the label for the phrase might not be present exactly in the caption) (named 'foreground'). The second list should have the single predominant background of the image to the foreground objects (named 'background'). If you do not find a phrase that corresponds to the background, return an empty list for the background. The third is an alternative background for the image, an alternative to the background you suggested earlier (named 'alt'). Here are some examples which also show the format in which you need to return the output. Please just return the JSON in the following format: \textbf{Exemplars}} $\cdots$ \textit{and here is the caption:}. We will provide the exemplars on our GitHub.
\vspace{5mm}

{\noindent \textbf{InstructPix2Pix.}}The prompt we use for InstructPix2Pix is: \textit{turn the background from original background to alternative background.} 
\subsection{Examples of top-\textit{k} identified by ASPIRE}

Table \ref{tab:top_k} shows the top-\textit{k} spuriously correlated features (or groups of features) for each class and for each dataset. As mentioned earlier, due to diversity in captions, the same kind of foreground object or background may be expressed with different phrases. ASPIRE thus returns groups of top-\textit{k} items rather than a single top-\textit{k} item for each \textit{k}.

\subsection{Collection of Test-Set for Hard ImageNet}

Our institution’s Institutional Review Board(IRB) has granted approval for the data collection. We followed the following steps for collecting a test set of the Hard ImageNet dataset:

\begin{enumerate}
    \item We first identified spurious features in the Hard ImageNet and verbalized them. These features were identified from annotations in the original proposed dataset by \citet{moayeri2022hard}.
    \item 3 annotators with extensive vision and language experience collected 1/3rd of the total 750 images. The annotators were not hired from any crowdsourcing platform and, in fact, were volunteers from our organization. The only instruction that was provided was that the image should have the primary target label of the image, and while majority group images should have the identified spurious features, minority group images should not.
    \item Post this step, each annotator validated the images collected by the other annotators.
    \item We filter the images for offensive content and replace them with non-offensive images, if any.
\end{enumerate}

\begin{table*}[t!]
    \centering
    \resizebox{2\columnwidth}{!}{
    \begin{tabular}{p{7cm}| c | c }
    \toprule
         \textbf{Dataset} & \textbf{Class} & \textbf{Top-\textit{k} groups} \\
         \midrule
          &  & \textcolor{ForestGreen}{\{}volleyball player, female volleyball player, two volleyball players\textcolor{ForestGreen}{\}} \\
          & Volleyball &  \textcolor{ForestGreen}{\{}woman, young girl, girl, women\textcolor{ForestGreen}{\}} \\
          &  &  \textcolor{blue}{\{}beach, sandy beach\textcolor{blue}{\}}\\
          
          \cline{2-3}
          
          &  & \textcolor{ForestGreen}{\{}keyboard, computer keyboard\textcolor{ForestGreen}{\}} \\
          & Spacebar & \textcolor{ForestGreen}{\{}number pad\textcolor{ForestGreen}{\}} \\
          &  &\textcolor{ForestGreen}{\{}mouse\textcolor{ForestGreen}{\}}\\
          
          \cline{2-3}
          
          &  & \textcolor{ForestGreen}{\{}girl, little girl, young girl, two girls\textcolor{ForestGreen}{\}} \\
          & Horizontal bar & \textcolor{ForestGreen}{\{}ballet, ballet barre\textcolor{ForestGreen}{\}} \\
          &  & \textcolor{blue}{\{}olympic games\textcolor{blue}{\}}\\
          
          \cline{2-3}
          
          &  & \textcolor{ForestGreen}{\{}boy, little boy, two boys\textcolor{ForestGreen}{\}} \\
          & Snorkel & \textcolor{ForestGreen}{\{}blue swimsuit, swimsuit\textcolor{ForestGreen}{\}} \\
          &  & \textcolor{blue}{\{}water, ocean\textcolor{blue}{\}}\\
          
          \cline{2-3}
          
          &  & \textcolor{ForestGreen}{\{}child, children, child's feet\textcolor{ForestGreen}{\}} \\
          & Balance Beam & \textcolor{ForestGreen}{\{}female gymnast, gymnast, gymnasts\textcolor{ForestGreen}{\}} \\
          &  &\textcolor{ForestGreen}{\{}split, leg split\textcolor{ForestGreen}{\}}\\
          
          \cline{2-3}
          
          &  & \textcolor{ForestGreen}{\{}back of the car, back of a car\textcolor{ForestGreen}{\}} \\
          & Seatbelt & \textcolor{ForestGreen}{\{}handle, door handle\textcolor{ForestGreen}{\}} \\
          &  & \textcolor{ForestGreen}{\{}seat, car seat, back seat\textcolor{ForestGreen}{\}}\\
          
          \cline{2-3}
          
          &  & \textcolor{ForestGreen}{\{}two dogs, dogs, dog, husky dogs, dog team\textcolor{ForestGreen}{\}} \\
          & Dog sled & \textcolor{blue}{\{}snowy hill, snowy landscape, snowy slope\textcolor{blue}{\}} \\
          &  & \textcolor{blue}{\{}snow\textcolor{blue}{\}}\\
          
          \cline{2-3}
          
          &  & \textcolor{ForestGreen}{\{}woman, young woman, women\textcolor{ForestGreen}{\}} \\
          \textbf{Hard ImageNet} & Miniskirt & \
          -\\
          &  & \-\\
          
          \cline{2-3}
          
          &  & \textcolor{ForestGreen}{\{}pink hat, hats\textcolor{ForestGreen}{\}}\\
          & Sunglasses & \textcolor{ForestGreen}{\{}woman, blonde woman, women\textcolor{ForestGreen}{\}}\\
          &  & \textcolor{ForestGreen}{\{}man, man's face\textcolor{ForestGreen}{\}}\\
          
          \cline{2-3}
          
          &  & \textcolor{ForestGreen}{\{}tree, tree branch, branch\textcolor{ForestGreen}{\}} \\
          & Howler monkey & \textcolor{ForestGreen}{\{}log, wooden bench, wooden beam\textcolor{ForestGreen}{\}}\\
          &  & \textcolor{ForestGreen}{\{}leaves\textcolor{ForestGreen}{\}}\\
          
          \cline{2-3}
          
          &  & \textcolor{ForestGreen}{\{}hockey player, hockey players, ice hockey players\textcolor{ForestGreen}{\}} \\
          & Puck & \textcolor{blue}{\{}ice, ice rink\textcolor{blue}{\}}\\
          &  & \textcolor{ForestGreen}{\{}hockey logo, hockey stick\textcolor{ForestGreen}{\}}\\
          
          \cline{2-3}
          
          &  & \textcolor{ForestGreen}{\{}boy, young boy, little boy\textcolor{ForestGreen}{\}}\\
          & Swimming cap & \textcolor{ForestGreen}{\{}pool, swimming pool, pool edge\textcolor{ForestGreen}{\}}\\
          &  & \textcolor{ForestGreen}{\{}swimmer, swimmers\textcolor{ForestGreen}{\}}\\
          
          \cline{2-3}
          
          &  & \textcolor{ForestGreen}{\{}chairs, chair, lawn chair\textcolor{ForestGreen}{\}}\\
          & Patio & \textcolor{ForestGreen}{\{}building, buildings\textcolor{ForestGreen}{\}} \\
          &  & \textcolor{ForestGreen}{\{}deck, new deck\textcolor{ForestGreen}{\}}\\
          
          \cline{2-3}
          
          &  & \textcolor{blue}{\{}mountain, mountain top, snowy mountain, snowy mountain side\textcolor{blue}{\}}\\
          & Ski & \textcolor{ForestGreen}{\{}ski poles, ski slope, ski lift\textcolor{ForestGreen}{\}}\\
          &  & \textcolor{ForestGreen}{\{}person, group of people\textcolor{ForestGreen}{\}}\\
          
          \cline{2-3}
          
          &  & \textcolor{blue}{\{}baseball field, field\textcolor{blue}{\}}\\
          & Baseball player & \textcolor{ForestGreen}{\{}baseball game, game\textcolor{ForestGreen}{\}}\\
          &  & \textcolor{blue}{\{}stadium\textcolor{blue}{\}}\\

         \hline
          &  & \textcolor{blue}{\{} lake, stream, pond\textcolor{blue}{\}}\\
          & Waterbird & \textcolor{blue}{\{}beach, sand\textcolor{blue}{\}}\\
          &  & \textcolor{blue}{\{}water, river bank\textcolor{blue}{\}}\\
          
          \cline{2-3}
          
          \textbf{Waterbirds} &  & \textcolor{blue}{\{}forest, bamboo forest\textcolor{blue}{\}}\\
          & Landbird & \textcolor{blue}{\{}woods, trees\textcolor{blue}{\}}\\
          &  & \textcolor{ForestGreen}{\{}branch, branches\textcolor{ForestGreen}{\}}\\
          
         \hline
          &  & \textcolor{blue}{\{}field, grass field, green field, grassy field, green grass covered field, lush green field\textcolor{blue}{\}} \\
          & Big dog & \textcolor{blue}{\{}ground, playground\textcolor{blue}{\}}\\
          &  & \textcolor{ForestGreen}{\{}floor, concrete floor\textcolor{ForestGreen}{\}}\\
          \cline{2-3}
          \textbf{Spuco Dogs} &  & \textcolor{ForestGreen}{\{}blanket, blue blanket, green blanket, red blanket, white blanket\textcolor{ForestGreen}{\}} \\
          & Small dog & \textcolor{ForestGreen}{\{}bed, dog bed, blue dog bed, small bed\textcolor{ForestGreen}{\}} \\
          &  & \textcolor{ForestGreen}{\{}couch, red couch, gray couch\textcolor{ForestGreen}{\}}\\

         \hline

          & Blonde & \textcolor{ForestGreen}{\{}woman, lady\textcolor{ForestGreen}{\}}\\
          \cline{2-3}
          \textbf{CelebA} & Not blonde & - \\

         \bottomrule
    \end{tabular}}
    \caption{Details of Top-\textit{k} (\textit{k}=3 for our experiments) spuriously correlated features per dataset and per class identified by ASPIRE. As discussed in the main paper, due to the fact that our captioning model generates diverse and variable phrases for the same type of object, we collapse these phrases into groups and instead work with groups (using the algorithm explained in Section 3 of the main paper) of spurious features. Groups in \textcolor{ForestGreen}{\{}\textcolor{ForestGreen}{\}} show spurious foreground objects while groups in \textcolor{blue}{\{}\textcolor{blue}{\}} are spurious backgrounds.}
    \label{tab:top_k}
\end{table*}

\section{GradCam Visualizations}
\label{sec:gradcam}

Figure \ref{fig:grid_of_images_1}  and \ref{fig:grid_of_images_2} illustrates the GradCAM visualizations of the features used by the last layer of a standard ERM model (ResNet-50), for prediction on the test set images, trained with and w/o ASPIRE augmentations on all 4 datasets used in our experiments. For a fair comparison and to clearly show the benefits of ASPIRE, we show GradCAM visualizations only for the standard ERM model, as all other baselines perform explicit steps to reduce reliance on such features. Standard ERM training is also still the most widely used methodology for training image classifiers.

While Fig. \ref{fig:grid_of_images_1} shows examples of the majority group images from the test set with spurious features, Figure \ref{fig:grid_of_images_2} shows examples of minority group images without any spurious features. For Fig. \ref{fig:grid_of_images_2} we show examples where the ERM classifier predicted the class of the image incorrectly (due to the absence of spurious features) while the one trained with ASPIRE predicted the class correctly. As we clearly see, in both cases, when trained with ASPIRE augmentations, the model learns to focus on core features rather than spurious ones while making predictions. 
% \vspace{-0.5em}
% \vspace{-0.5em}
\begin{figure}[t]
    \centering
    \begin{subfigure}{0.2\textwidth}
        \includegraphics[width=0.9\linewidth]{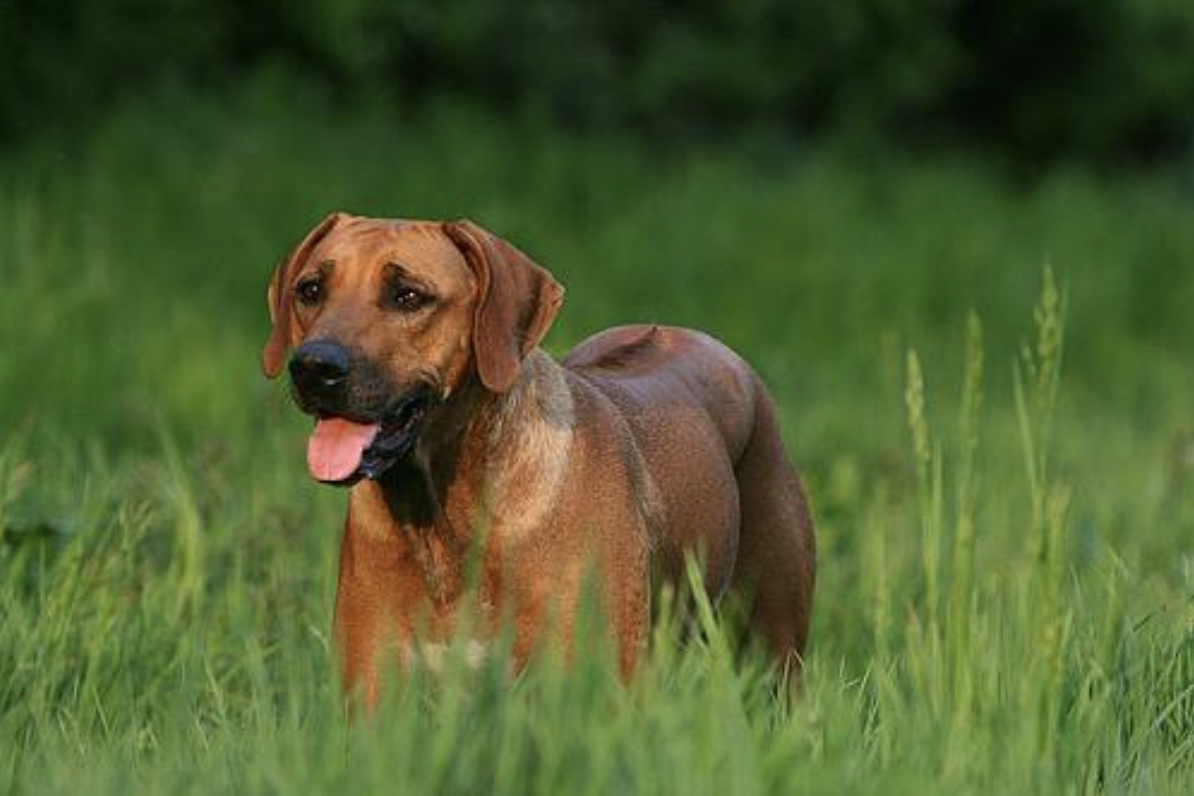}
        \caption{Original Image}
        \label{fig:dog_orig}
    \end{subfigure}
    \hfill
    \begin{subfigure}{0.2\textwidth}
        \includegraphics[width=0.9\linewidth]{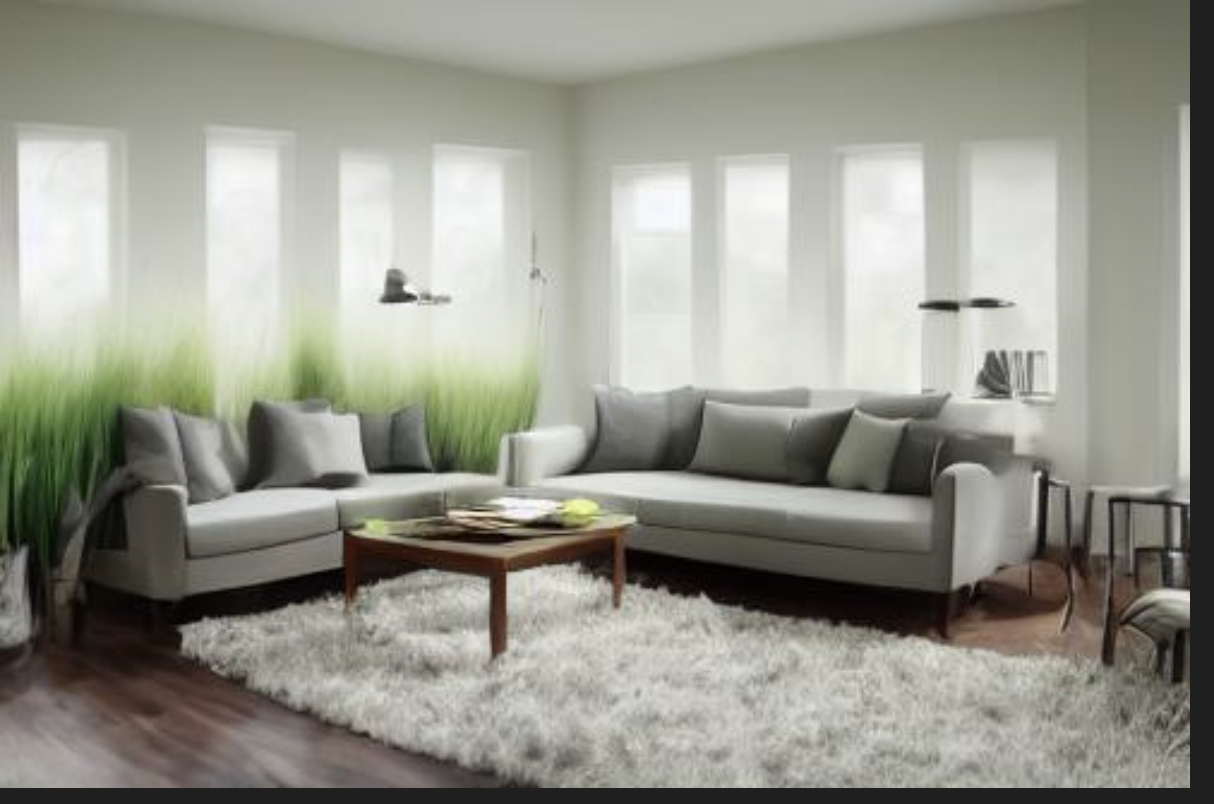}
        \caption{Edited Image}
        \label{fig:dog_edit}
    \end{subfigure}

    \begin{subfigure}{0.2\textwidth}
        \includegraphics[width=0.9\linewidth]{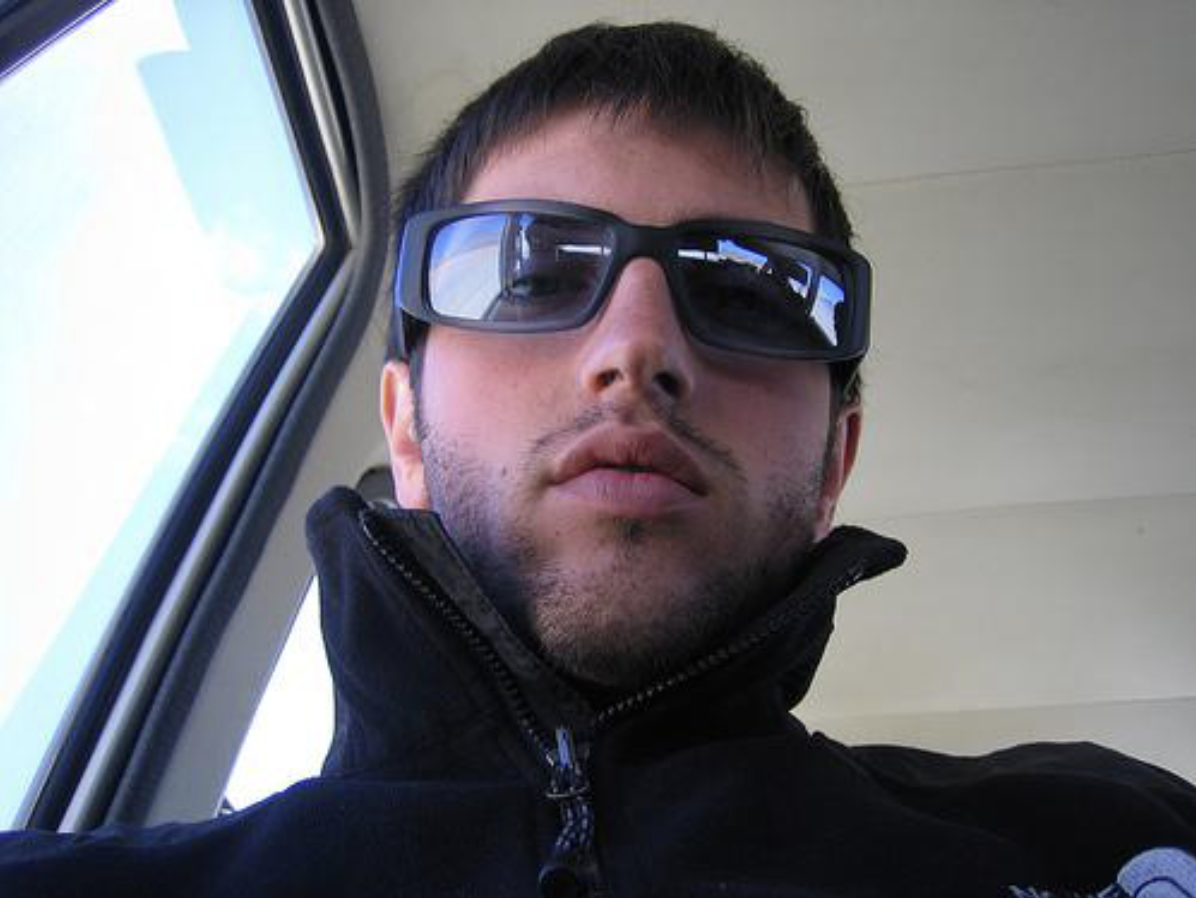}
        \caption{Original Image}
        \label{fig:person_orig}
    \end{subfigure}
    \hfill
    \begin{subfigure}{0.2\textwidth}
        \includegraphics[width=0.9\linewidth]{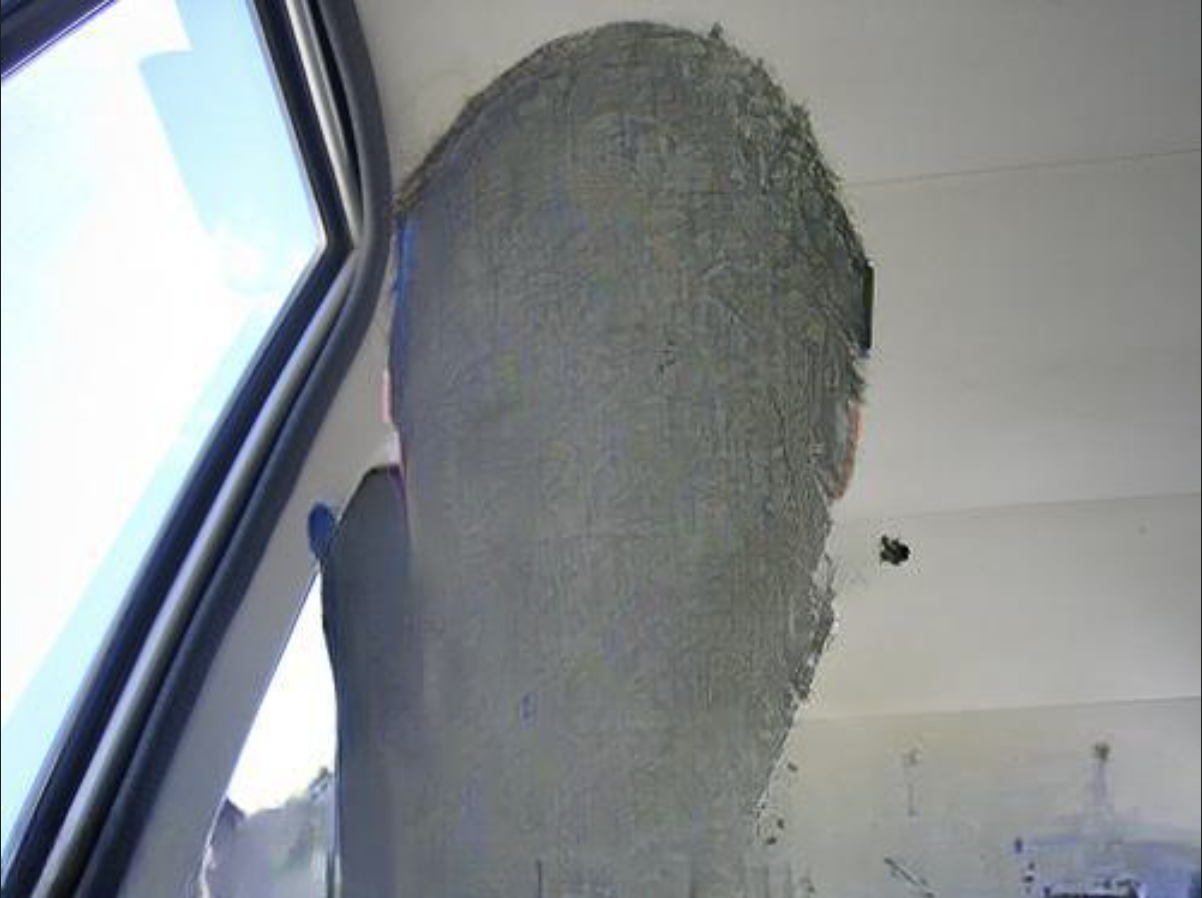}
        \caption{Edited Image}
        \label{fig:person_edit}
    \end{subfigure}
    \vspace{-0.75em}
    \caption{\small Images illustrating cases of ASPIRE failures.}
    \vspace{-1em}
    \label{fig:failure_cases}
\end{figure}
\section{Generation Examples}
\label{sec:generation}

Table \ref{fig:example_images} shows examples of original images from the train set, the edited images from the ASPIRE pipeline and augmentations generated using ASPIRE. To the left of the generated augmentations, we also mention the top-\textit{k} spurious correlations discovered by ASPIRE for the particular class. ASPIRE generates diverse augmentations with the desired non-spurious features that can be used to train robust models

\section{ASPIRE Failure Cases}
\label{sec:failure}

This section lists some failure cases of our proposed ASPIRE framework. As APSIRE leverages external models in its pipeline, the success of ASPIRE at times depends on the capabilities of these models. For generating augmentations, we notice the following failure cases:

\begin{enumerate}
    \item \textbf{Superimposition of other foreground objects on the foreground object of interest.} Recall that ASPIRE detects foreground objects to remove (for spurious correlation detection) by parsing captions. These objects are then removed to detect if the object is spurious or not. In cases where another foreground object in the image is superimposing the foreground object, though the language-grounded pipeline is able to detect it properly, the inpainting model is at times unable to precisely remove just that object without not removing the superimposing foreground object and removes both the original object and the object superimposing it. An example is a \textit{human} wearing \textit{spectacles}, where we only want to remove the human, but the inpainting model removes both the human and the spectacles it is wearing. We provide an example of this case in Figure~\ref{fig:failure_cases} (bottom row).

    \item \textbf{Foreground objects change on changing background.} InstructPix2Pix, at times, tends to change the foreground object when prompted to change the background significantly, for example, changing \textit{outdoor background} $\rightarrow$ \textit{indoor background}. We provide an example of this case in Figure~\ref{fig:failure_cases} (top row).

    \item \textbf{Bias in Stable Diffusion.} Although our Stable Diffusion fine-tuning step, with textual inversion, helps overcome its current biases, limited samples present for fine-tuning sometimes hurt this adaptation. For example, even after fine-tuning, Stable Diffusion might generate images of \textit{dog sled} with dogs in it.

\end{enumerate}

\section{Additional Details}
\subsection{Model Parameters}
Git-Large-Coco has $\approx$ 300M parameters with a CLIP/ViT-L/14 image encoder and a 6 layer transformer decoder with 12 attention heads and 768 hidden-state. Stable Diffusion is a $\approx$ 860M parameter UNet and $\approx$ 123M parameter text encoder model. ResNet-50 has $\approx$ 25M parameters with 50 layers. 

\subsection{Compute Infrastructure}
All our experiments are conducted on NVIDIA A100 GPUs. We batch prompted LLaMa-2 13B, with a BS of 16, where LLaMa-2 performed distributed inference on 4 A100 GPUs. That translates to 52.51 TFLOPs per batch. Fine-tuning SD with textual inversion with a BS of 8 takes and an avg. of $\approx$1 hour. For generating 1$\times$ augmentation, we use 1 A100 GPU for an average $\approx$1.2 hours in total.

\subsection{Implementation Software and Packages}
 We implement all our models in PyTorch\footnote{\href{https://pytorch.org}{https://pytorch.org/}} and use the HuggingFace\footnote{\href{https://huggingface.co/}{https://huggingface.co/}} implementations of ERM, Git~\cite{wang2022git}, LLaMa-2 13B~\cite{touvron2023llama} and InstructPix2Pix \cite{brooks2023instructpix2pix}. We also use the following code/components in our pipeline Grounding DINO\footnote{\href{https://github.com/IDEA-Research/GroundingDINO}{https://github.com/IDEA-Research/GroundingDINO}} \cite{liu2023grounding}, Segment Anything\footnote{\href{https://github.com/facebookresearch/segment-anything}{https://github.com/facebookresearch/segment-anything}} \cite{kirillov2023segment} and Stable Diffusion using textual-inversion\footnote{\href{https://github.com/rinongal/textual_inversion}{https://github.com/rinongal/textual\_inversion}} \cite{gal2023an}. \\

We also use the following repositories for running the baselines: Group DRO\footnote{\href{https://github.com/kohpangwei/group_DRO}{https://github.com/kohpangwei/group\_DRO}} \cite{sagawa2019distributionally}, SUBG\footnote{\href{https://github.com/facebookresearch/BalancingGroups}{https://github.com/facebookresearch/BalancingGroups}} \cite{idrissi2022simple}, JTT\footnote{\href{https://github.com/anniesch/jtt}{https://github.com/anniesch/jtt}} \cite{liu2021just}, Learning from Failure \footnote{\href{https://github.com/alinlab/LfF}{https://github.com/alinlab/LfF}} \cite{nam2020learning}, Correct-n-Contrast\footnote{\href{https://github.com/HazyResearch/correct-n-contrast}{https://github.com/HazyResearch/correct-n-contrast}} \cite{zhang2022correct}, Deep Feature Reweighting\footnote{\href{https://github.com/PolinaKirichenko/deep_feature_reweighting}{https://github.com/PolinaKirichenko/deep\_feature\_reweighting}} \cite{kirichenko2023last}, MaskTune\footnote{\href{https://github.com/aliasgharkhani/masktune}{https://github.com/aliasgharkhani/masktune}} \cite{asgari2022masktune} and DivDis\footnote{\href{https://github.com/yoonholee/DivDis}{https://github.com/yoonholee/DivDis}} \cite{DBLP:journals/corr/abs-2202-03418}.

All the above GitHub code has been released under an MIT license, free for research use.
\begin{figure*}[t]
    \centering
    \begin{subfigure}{0.2\linewidth}
        \includegraphics[width=\linewidth]        {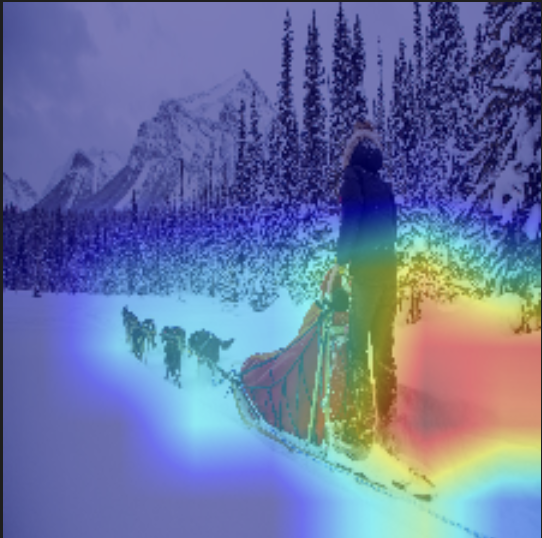}
        \caption{Dog sled w/o ASPIRE}
    \end{subfigure}
    \hfill
    \begin{subfigure}{0.2\linewidth}
        \includegraphics[width=\linewidth]{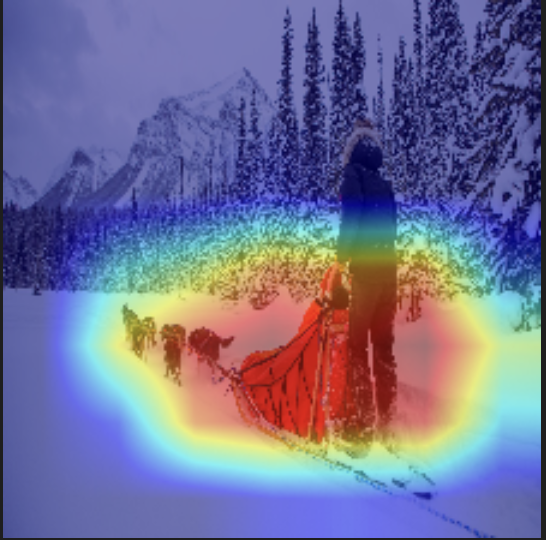}
        \caption{Dog sled w/ ASPIRE}
    \end{subfigure}
    \hfill
    \begin{subfigure}{0.2\linewidth}
        \includegraphics[width=\linewidth]{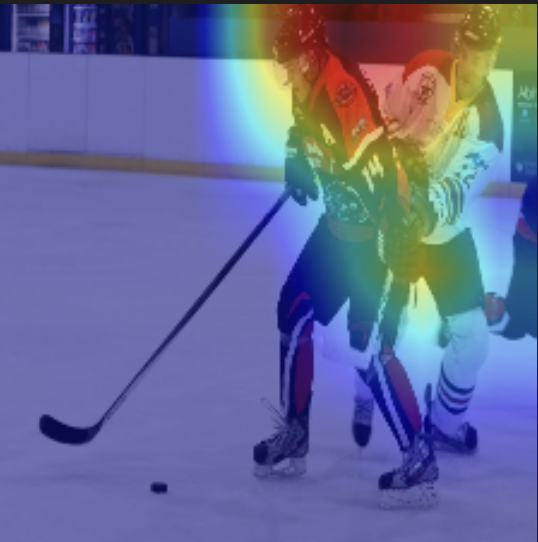}
        \caption{Puck w/o ASPIRE}
    \end{subfigure}
    \hfill
    \begin{subfigure}{0.2\linewidth}
        \includegraphics[width=\linewidth]{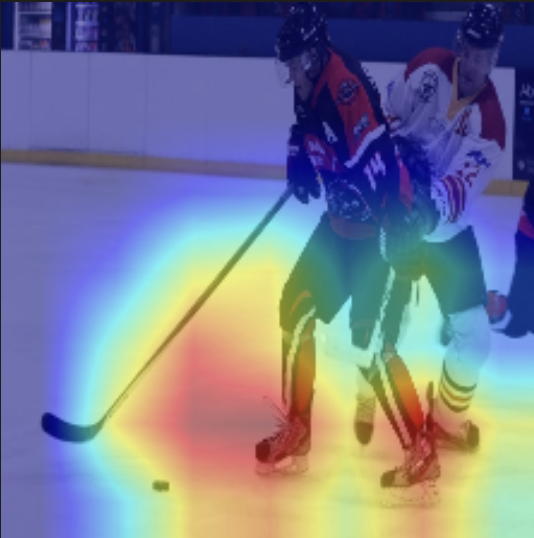}
        \caption{Puck w/ ASPIRE}
    \end{subfigure}
    
    \medskip
    
    \begin{subfigure}{0.2\linewidth}
        \includegraphics[width=\linewidth]{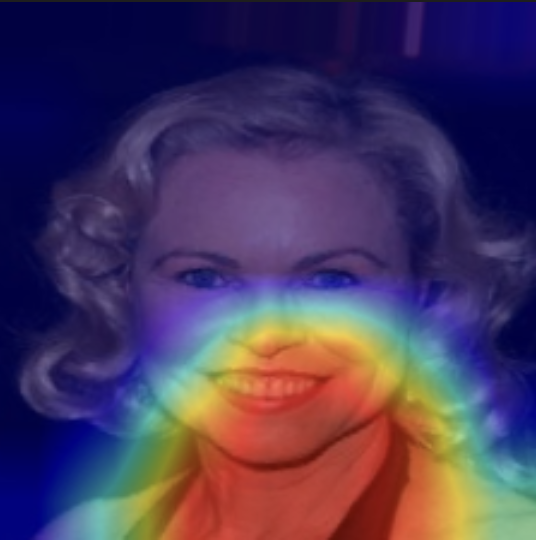}
        \caption{Blonde female w/o ASPIRE}
    \end{subfigure}
    \hfill
    \begin{subfigure}{0.2\linewidth}
        \includegraphics[width=\linewidth]{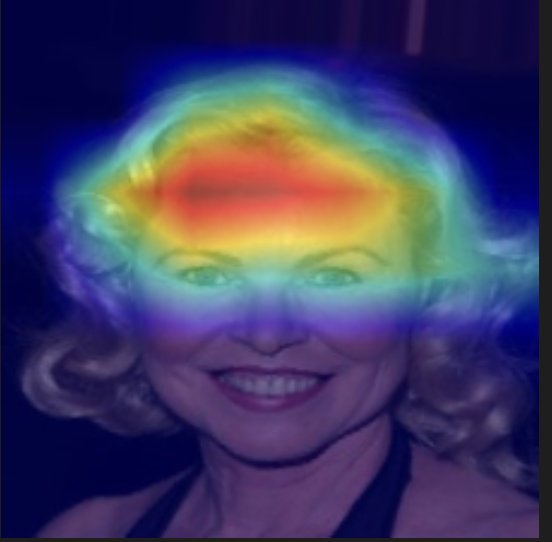}
        \caption{Blonde female w/ ASPIRE}
    \end{subfigure}
    \hfill
    \begin{subfigure}{0.2\linewidth}
        \includegraphics[width=\linewidth]{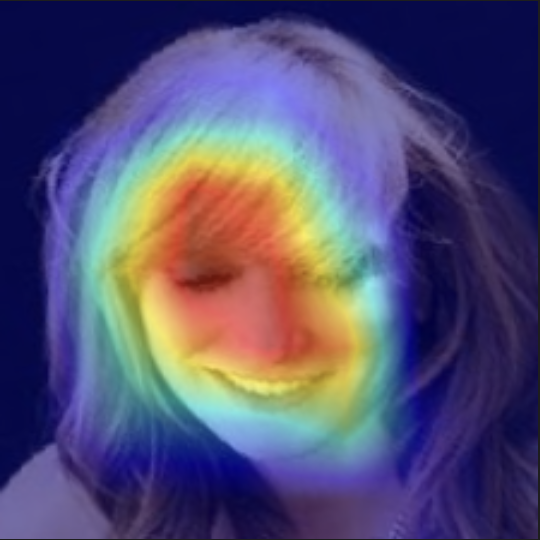}
        \caption{Blonde female w/o ASPIRE}
    \end{subfigure}
    \hfill
    \begin{subfigure}{0.2\linewidth}
        \includegraphics[width=\linewidth]{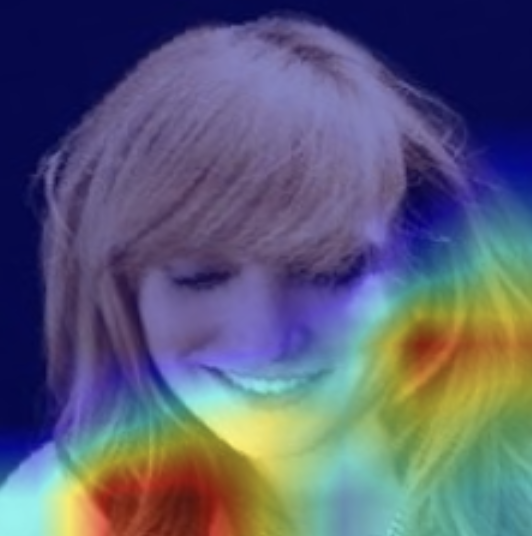}
        \caption{Blonde female w/ ASPIRE}
    \end{subfigure}

    \medskip
    
    \begin{subfigure}{0.2\linewidth}
        \includegraphics[width=\linewidth]{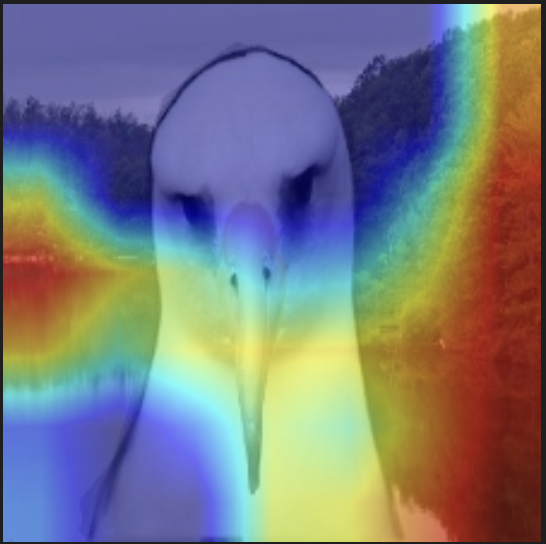}
        \caption{Waterbird w/o ASPIRE}
    \end{subfigure}
    \hfill
    \begin{subfigure}{0.2\linewidth}
        \includegraphics[width=\linewidth]{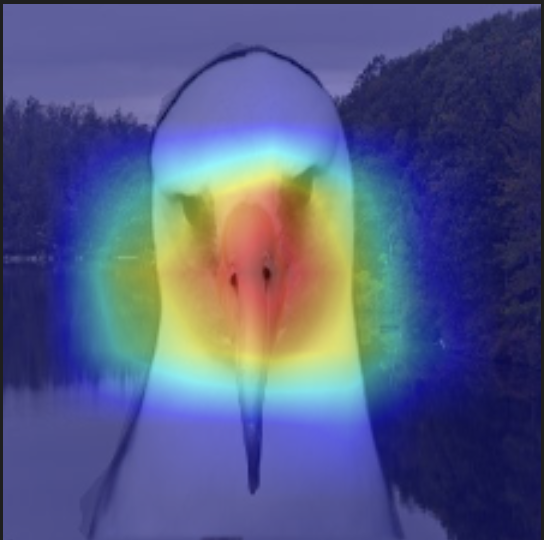}
        \caption{Waterbird w/ ASPIRE}
    \end{subfigure}
    \hfill
    \begin{subfigure}{0.2\linewidth}
        \includegraphics[width=\linewidth]{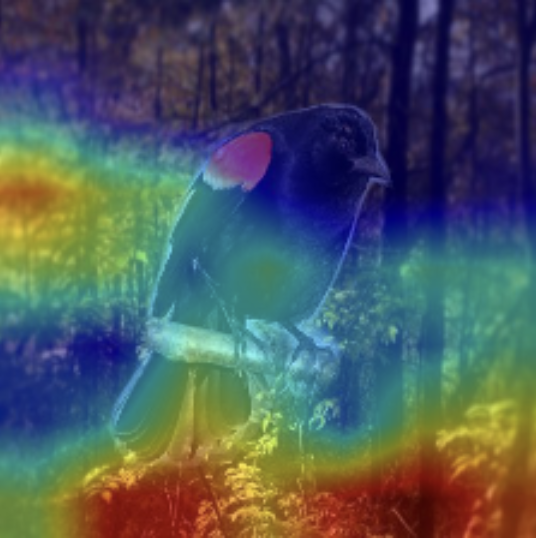}
        \caption{Landbird w/o ASPIRE}
    \end{subfigure}
    \hfill
    \begin{subfigure}{0.2\linewidth}
        \includegraphics[width=\linewidth]{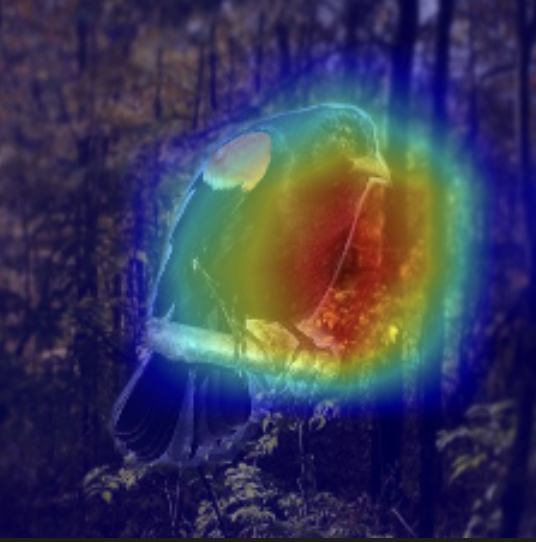}
        \caption{Landbird w/ ASPIRE}
    \end{subfigure}

    \medskip
    
    \begin{subfigure}{0.2\linewidth}
        \includegraphics[width=\linewidth]{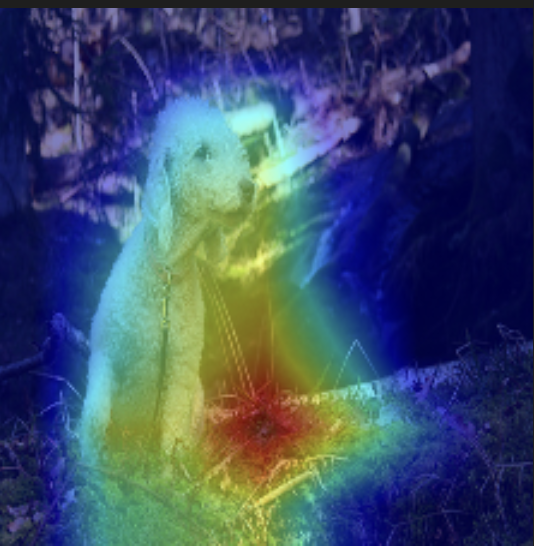}
        \caption{Bigdog w/o ASPIRE}
    \end{subfigure}
    \hfill
    \begin{subfigure}{0.2\linewidth}
        \includegraphics[width=\linewidth]{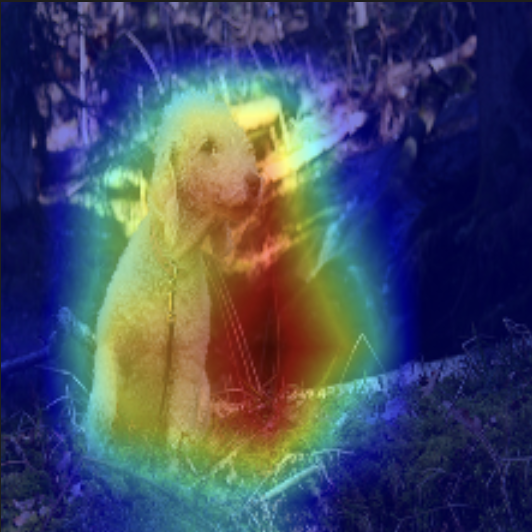}
        \caption{Bigdog w/ ASPIRE}
    \end{subfigure}
    \hfill
    \begin{subfigure}{0.2\linewidth}
        \includegraphics[width=\linewidth]{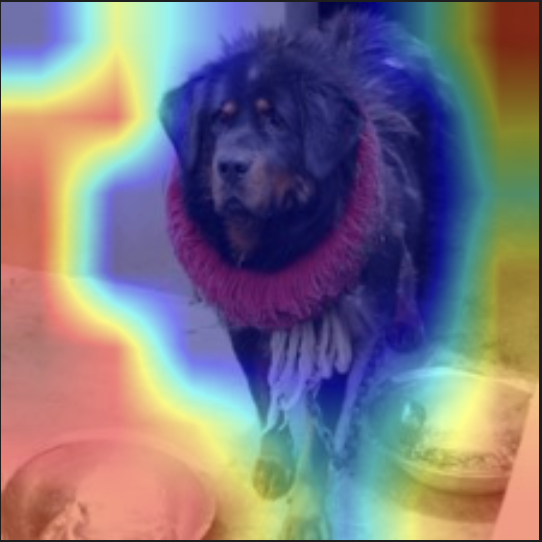}
        \caption{Smalldog w/o ASPIRE}
    \end{subfigure}
    \hfill
    \begin{subfigure}{0.2\linewidth}
        \includegraphics[width=\linewidth]{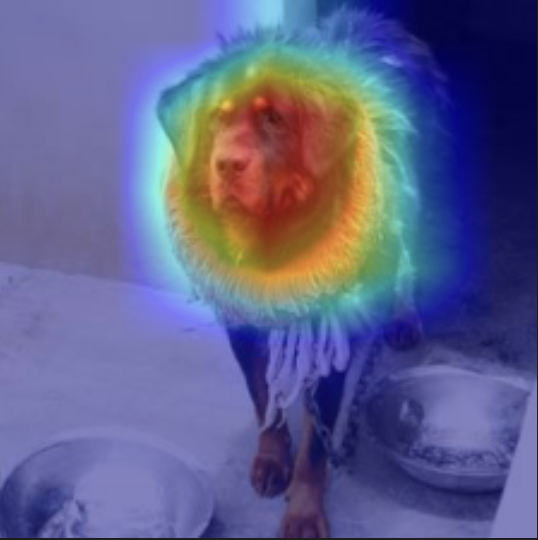}
        \caption{Smalldog w/ ASPIRE}
    \end{subfigure}
    
    \caption{GradCam visualizations of features used by the last layer of a standard ERM model to predict majority group images (with spuriously correlated features) from the test set of 4 datasets.}
    \label{fig:grid_of_images_1}
\end{figure*}

\begin{figure*}[t!]
    \centering
    \begin{subfigure}{0.2\linewidth}
        \includegraphics[width=\linewidth]{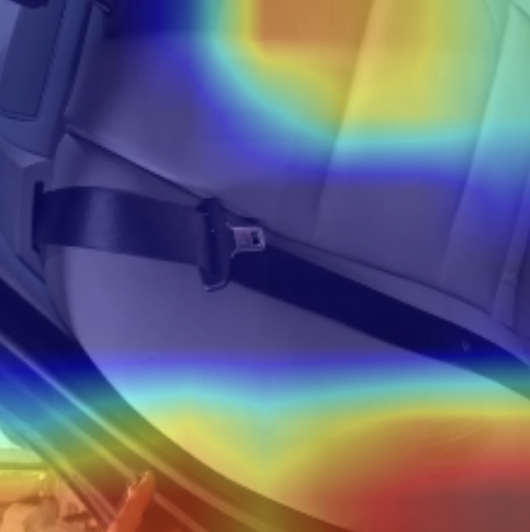}
        \caption{Seatbelt w/o ASPIRE}
    \end{subfigure}
    \hfill
    \begin{subfigure}{0.2\linewidth}
        \includegraphics[width=\linewidth]{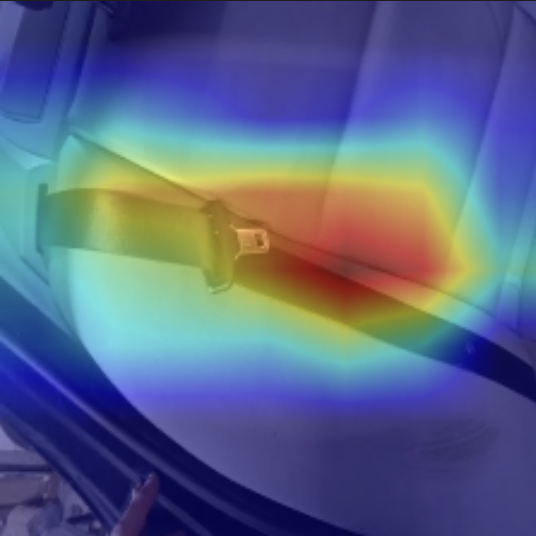}
        \caption{Seatbelt w/ ASPIRE}
    \end{subfigure}
    \hfill
    \begin{subfigure}{0.2\linewidth}
        \includegraphics[width=\linewidth]{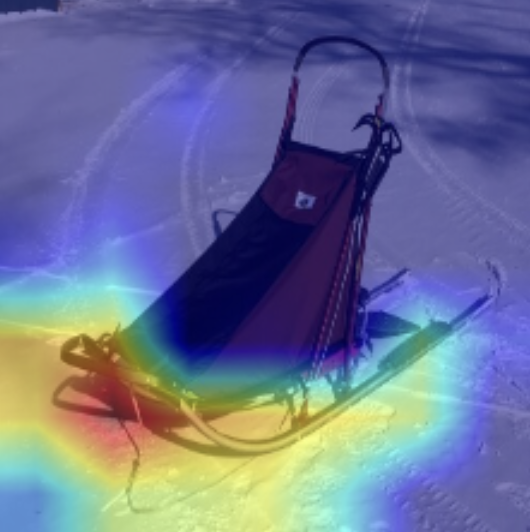}
        \caption{Dogsled w/o ASPIRE}
    \end{subfigure}
    \hfill
    \begin{subfigure}{0.2\linewidth}
        \includegraphics[width=\linewidth]{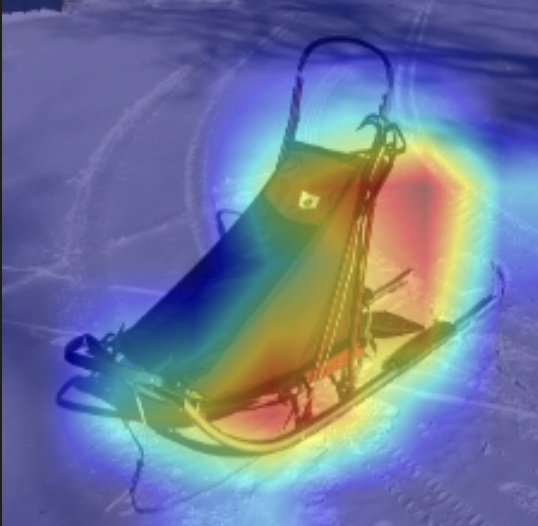}
        \caption{Dogsled w/ ASPIRE}
    \end{subfigure}
    
    \medskip
    
    \begin{subfigure}{0.2\linewidth}
        \includegraphics[width=\linewidth]{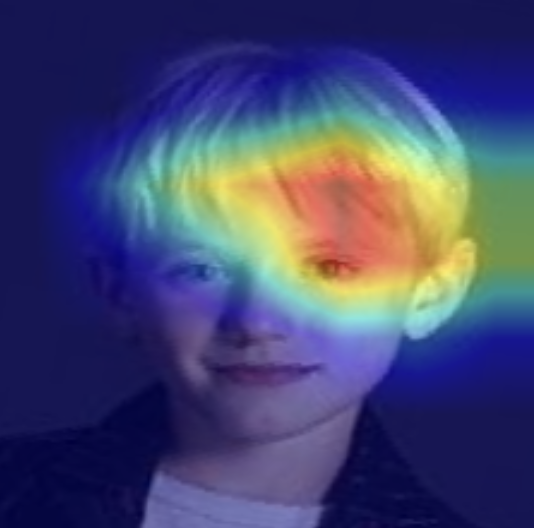}
        \caption{Male blonde w/o ASPIRE}
    \end{subfigure}
    \hfill
    \begin{subfigure}{0.2\linewidth}
        \includegraphics[width=\linewidth]{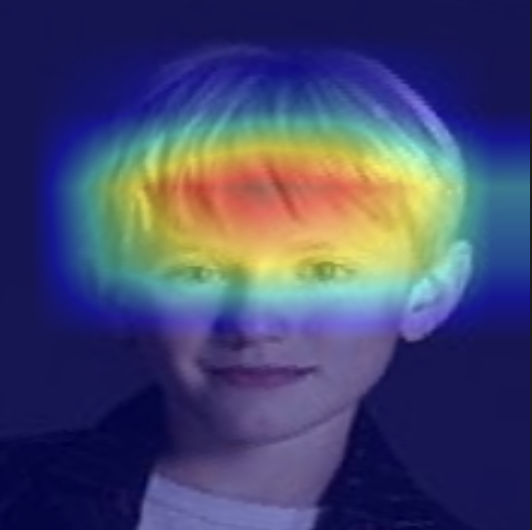}
        \caption{Male blonde w/ ASPIRE}
    \end{subfigure}
    \hfill
    \begin{subfigure}{0.2\linewidth}
        \includegraphics[width=\linewidth]{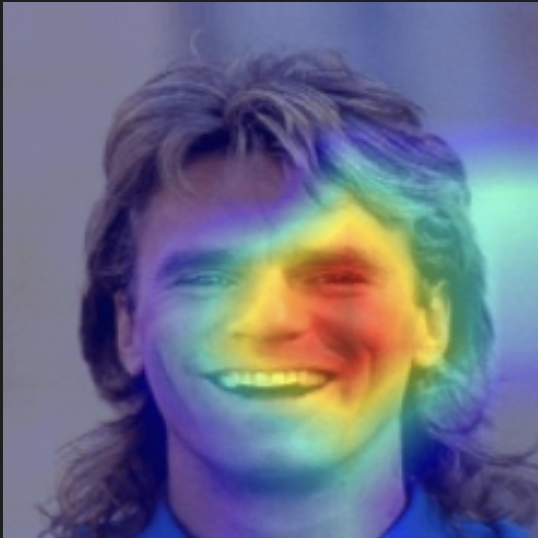}
        \caption{Male blonde w/o ASPIRE}
    \end{subfigure}
    \hfill
    \begin{subfigure}{0.2\linewidth}
        \includegraphics[width=\linewidth]{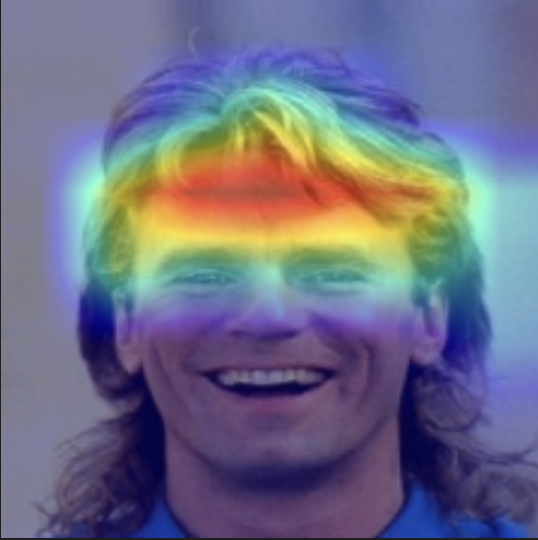}
        \caption{Male blonde w/ ASPIRE}
    \end{subfigure}

    \medskip
    
    \begin{subfigure}{0.2\linewidth}
        \includegraphics[width=\linewidth]{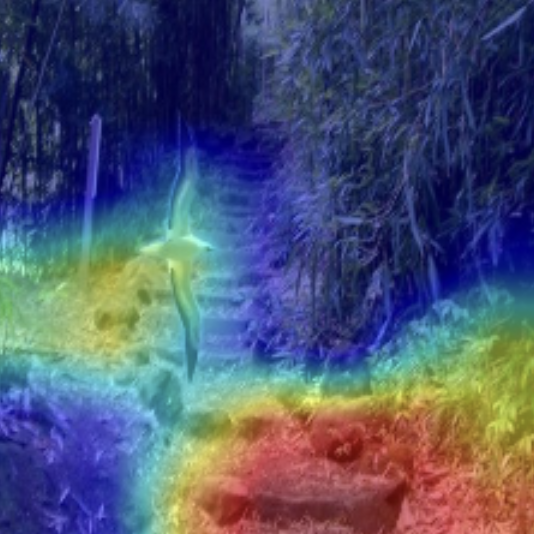}
        \caption{Waterbird w/o ASPIRE}
    \end{subfigure}
    \hfill
    \begin{subfigure}{0.2\linewidth}
        \includegraphics[width=\linewidth]{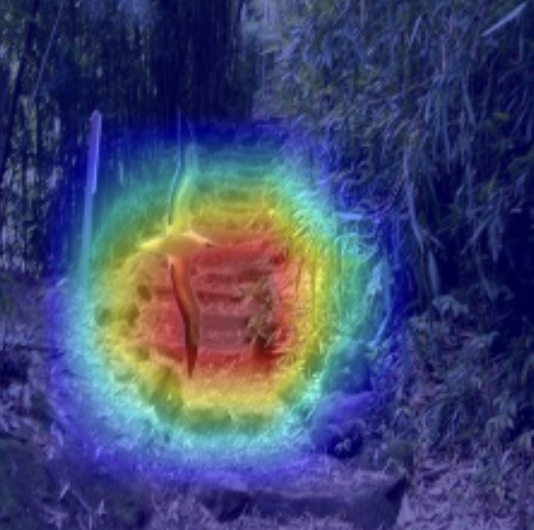}
        \caption{Waterbird w/ ASPIRE}
    \end{subfigure}
    \hfill
    \begin{subfigure}{0.2\linewidth}
        \includegraphics[width=\linewidth]{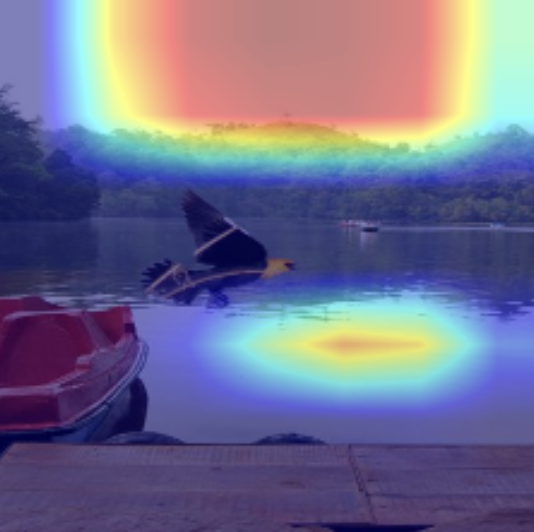}
        \caption{Landbird w/ ASPIRE}
    \end{subfigure}
    \hfill
    \begin{subfigure}{0.2\linewidth}
        \includegraphics[width=\linewidth]{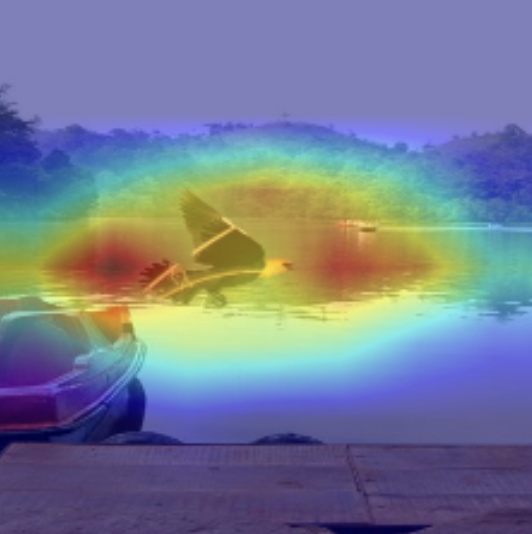}
        \caption{Landbird w/o ASPIRE}
    \end{subfigure}

    \medskip
    
    \begin{subfigure}{0.2\linewidth}
        \includegraphics[width=\linewidth]{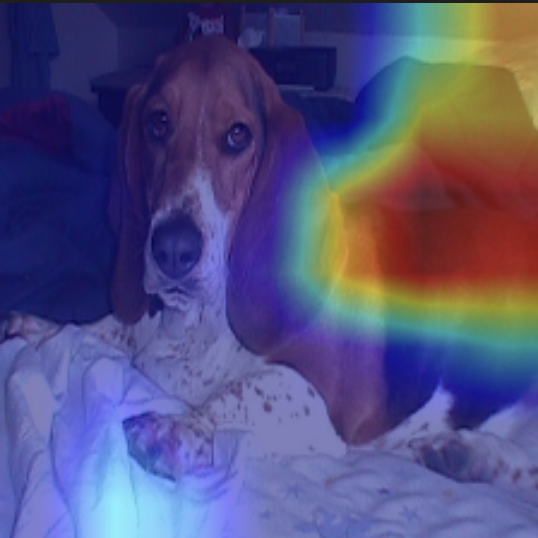}
        \caption{Bigdog w/o ASPIRE}
    \end{subfigure}
    \hfill
    \begin{subfigure}{0.2\linewidth}
        \includegraphics[width=\linewidth]{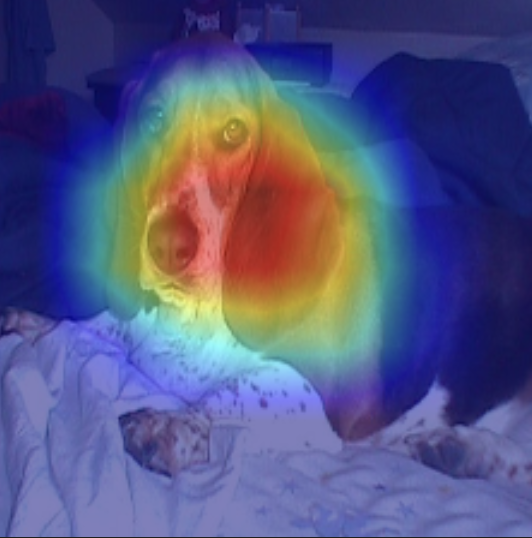}
        \caption{Bigdog w/ ASPIRE}
    \end{subfigure}
    \hfill
    \begin{subfigure}{0.2\linewidth}
        \includegraphics[width=\linewidth]{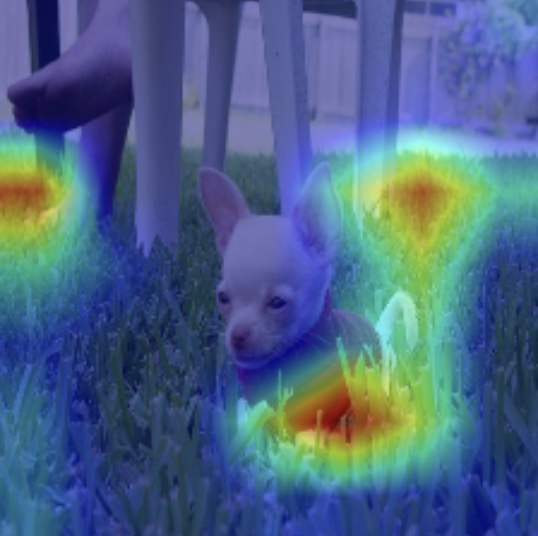}
        \caption{Smalldog w/o ASPIRE}
    \end{subfigure}
    \hfill
    \begin{subfigure}{0.2\linewidth}
        \includegraphics[width=\linewidth]{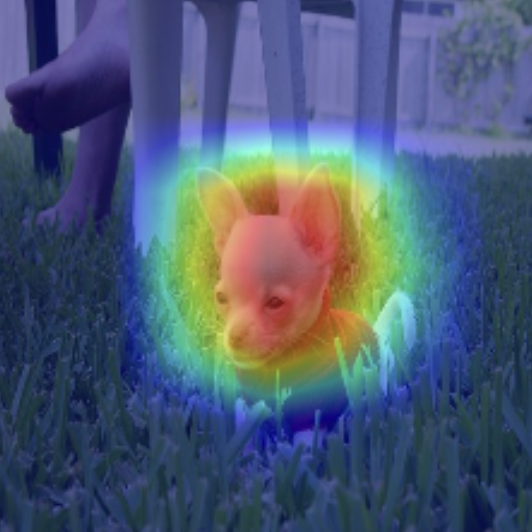}
        \caption{Smalldog w/ ASPIRE}
    \end{subfigure}
    
    \caption{GradCam visualizations of features used by the last layer of a standard ERM model to predict majority group images (\textit{without} spuriously correlated features) from the test set of 4 datasets.}
    \label{fig:grid_of_images_2}
\end{figure*}
\subsection{Dataset Links}

We use the Waterbirds~\footnote{https://www.vision.caltech.edu/visipedia/CUB-200.html}, SPUCO Dogs~\footnote{https://github.com/BigML-CS-UCLA/SpuCo}, Hard ImageNet~\footnote{https://openreview.net/forum?id=76w7bsdViZf} and the CelebA~\footnote{https://mmlab.ie.cuhk.edu.hk/projects/CelebA.html} dataset. All the datasets are free for research use.

\subsection{Potential Risks}
Generative models are prone to hallucinate and potentially generate nonsensical, unfaithful or harmful content to the provided source input that it is conditioned on.

\begin{figure*}[t!]
\includegraphics[width=\textwidth,height=0.9\textheight]{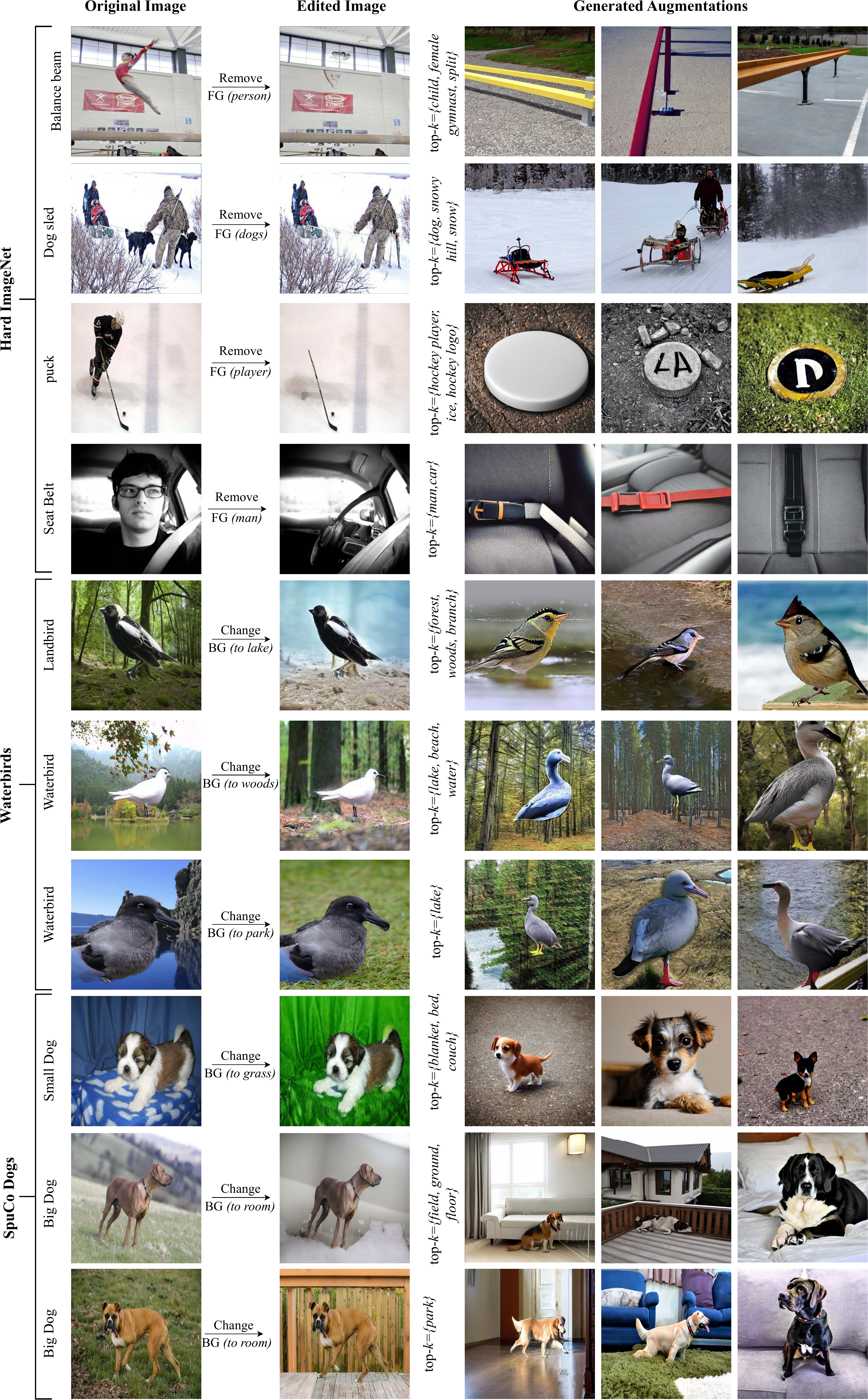}
    \caption{Examples of \textbf{Original Images}, \textbf{Edited Images} from the ASPIRE pipeline and \textbf{Generated Augmentations}. To the left of the \textbf{Generated Augmentations}, we also mention the top-\textit{k} spurious correlations discovered by ASPIRE for the particular class. ASPIRE generates diverse augmentations with the desired non-spurious features that can be used to train robust models.}
    \label{fig:example_images}
\end{figure*}

\end{document}